\documentclass[10pt,twocolumn,letterpaper]{article}

\usepackage{cvpr}
\usepackage{times}
\usepackage{epsfig}
\usepackage{graphicx}
\usepackage{amsmath}
\usepackage{amssymb}
\usepackage{subfig}
\usepackage{booktabs}  
\usepackage{multirow}

\usepackage[pagebackref=true,breaklinks=true,letterpaper=true,colorlinks,bookmarks=false]{hyperref}

 \cvprfinalcopy 

\def\cvprPaperID{483} 

\ifcvprfinal\fi
\begin{document}

\title{Left-Right  Comparative Recurrent Model for Stereo Matching}

\author{Zequn Jie$^{*}$ \quad Pengfei Wang$^\dag$ \quad Yonggen Ling$^*$ \quad Bo Zhao$^\mathsection$ \quad Yunchao Wei$^\wr$ \quad Jiashi Feng$^\dag$ \quad Wei Liu$^*$ \\ \\
	$^*$Tencent AI Lab \quad  $^\dag$National University of Singapore \\ $^\mathsection$University of British  Columbia \quad $^\wr$University of Illinois Urbana-Champaign \\ {\tt\small \{zequn.nus, wpfhtl, lingyg2008, zhaobo.cs,  wychao1987, jshfeng\}@gmail.com \quad wliu@ee.columbia.edu}
}
\maketitle
\thispagestyle{empty}

\begin{abstract}
	Leveraging the disparity information from both  left and right views is crucial for stereo disparity estimation. Left-right consistency check is an effective way to enhance the disparity estimation by referring to the information from the opposite view. However, the conventional left-right consistency check is an isolated post-processing step and heavily hand-crafted. This paper proposes a novel left-right comparative recurrent model to perform left-right consistency checking jointly with   disparity estimation. At each recurrent step, the model produces disparity results for both views, and then performs online left-right comparison to identify the mismatched regions which may probably contain erroneously labeled pixels. A soft attention mechanism is introduced, which employs the learned error maps for better guiding the model to selectively focus on refining the unreliable regions at the next recurrent step. In this way, the generated disparity maps are progressively improved by the proposed recurrent model. Extensive evaluations on  KITTI 2015, Scene Flow and Middlebury benchmarks validate the effectiveness of our model,  demonstrating that state-of-the-art stereo disparity estimation results can be achieved by this new model.
\end{abstract}

\vspace{-0.5cm}
\section{Introduction}

This paper aims at the problem of computing the dense disparity map between a rectified stereo pair of images. Stereo disparity estimation is   core to many computer vision applications, including robotics and autonomous vehicles  \cite{geiger2013vision, Geiger2012CVPR, maddern2016real}.  Finding local correspondences between two images plays a key role in generating high-quality disparity maps.  Modern   stereo matching methods rely on deep neural networks to learn powerful visual representations and compute  more accurate  matching costs between  left and right views.

\begin{figure}
	\captionsetup[subfigure]{labelformat=empty}
	\centering
	\subfloat[]{\includegraphics[width=7cm,height=1.5cm]{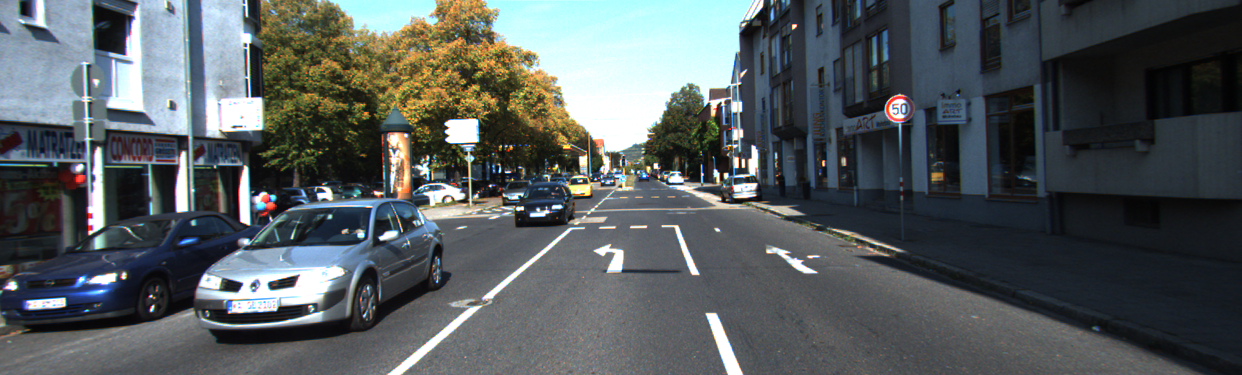}		}
	\\
	\vspace{-0.75cm}
	\subfloat[]{\includegraphics[width=7cm,height=1.5cm]{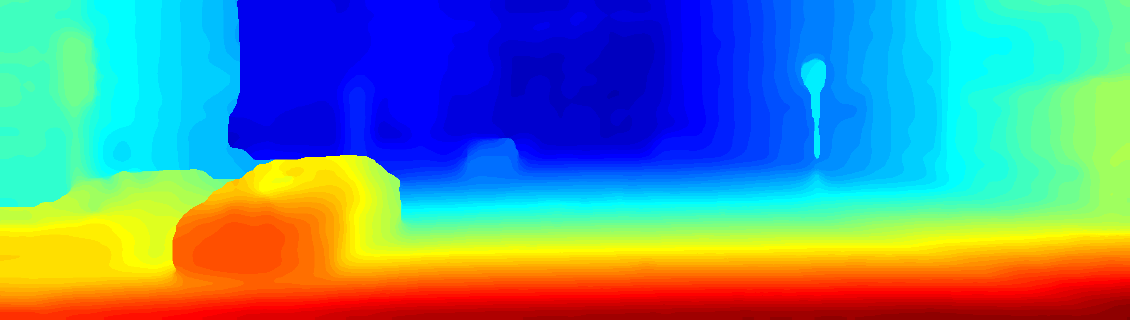}}
	\\
	\vspace{-0.75cm}
	\subfloat[]{\includegraphics[width=7cm,height=1.5cm]{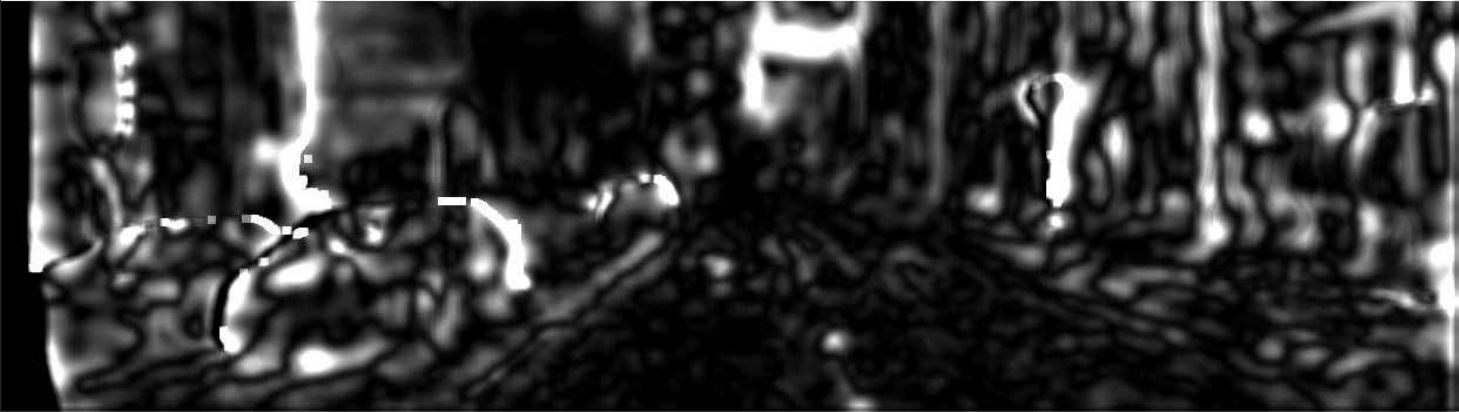}
	}
	\\
	\vspace{-0.75cm}
	\subfloat[]{\includegraphics[width=7cm,height=1.5cm]{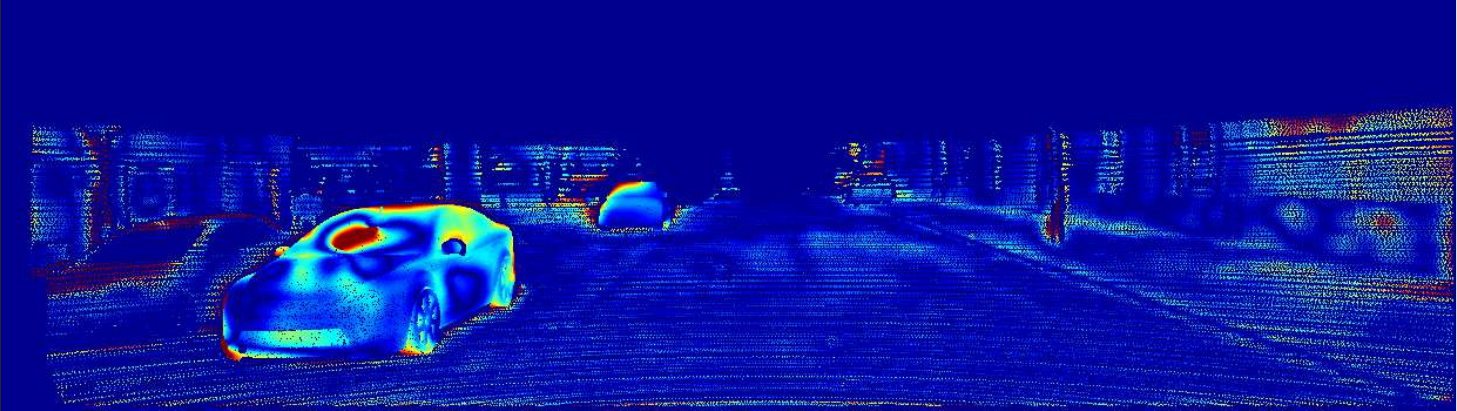}
	}
	\\
	\vspace{-0.5cm}
	\caption{From top to bottom: the input image,  predicted disparity map, learned attention map showing the potential erroneously labeled regions by the LRCR model, and the real error map (difference between the predicted and ground-truth maps).}
	\vspace{-0.7cm}
	\label{fig:fig1}
	
\end{figure}

However, it is still challenging for the current methods to deal with ill-posed regions, such as object occlusions, reflective regions, and repetitive patterns.  It is also observed that the mismatched pixels between the left and right views usually appear in the error-prone regions, including occluded objects, textureless areas, and sophisticated image borders (see Fig.~\ref{fig:fig1}).   Taking advantage of disparity information from  both the views to verify the left-right mutual consistency  is an effective strategy to identify the unreliable regions. By doing this,  the stereo disparity estimation results can be selectively improved by being refined on the mismatched regions. Traditional left-right consistency check is performed only as an offline post-processing step after the disparity map generation. Moreover, it is highly hand-crafted and hardwired\textemdash it only refines the pixels having errors   above a   mismatching threshold by interpolating their fixed local neighbors. The results are thus fragile due to low-quality local features and potential errors in the matching cost computation.  As such,  the traditional pipelines, whose regularization steps involve hand-engineered, shallow cost aggregation and disparity optimization (SGM or MRF), have been proven to be inferior \cite{seki2016patch} and sub-optimal for  stereo disparity estimation. 

To overcome the above issue, we propose a novel \textit{Left-Right Comparative Recurrent} (LRCR) model to integrate the left-right consistency check and disparity generation into a unified pipeline, which allows them to mutually boost and effectively resolve the drawbacks of   offline  consistency check. Formulated as a recurrent neural network (RNN)-style model, the LRCR model can learn to progressively improve the disparity estimation for both the left and right images by exploiting  the learned left-right consistency. In this way, both  disparity maps for two views favorably converge to stable and accurate predictions eventually. LRCR creatively introduces an  attention mechanism accompanying recurrent learning to  simultaneously check consistency  and select proper regions for refinement. Concretely,  at each recurrent step, the model processes both views in parallel, and produces both disparity maps as well as their associated error maps by performing the online left-right comparison. The error maps reflect the correctness of the obtained disparities by ``referring'' the disparity map of the other view. Treating the error maps as  soft attention maps, the model is guided to concentrate more on the potential erroneously predicted regions in the next recurrent step. Such an error-diagnosis self-improving scheme allows the model to automatically improve the estimation of both views without tedious extra supervision over  the erroneously labeled regions. Moreover, incorporating the left-right consistency into the model training achieves the desirable consistency between the training and inference (application) phases. Thus, LRCR can improve the disparity estimation performance in a more straightforward and expected manner, with better optimization quality.

The proposed left-right comparative recurrent model is based on a convolutional Long-Short Term Memory (ConvLSTM) network, considering its superior capability of  incorporating contextual information from multiple disparities. Besides, the historical disparity estimation stored in LSTM memory cells   automatically flows to following steps, and provides a reasonably good initial disparity estimation to facilitate  generating  higher-quality disparities in later steps. The proposed LRCR model replaces the conventional hand-crafted regularization methods \cite{scharstein2002taxonomy} plus the ``Winner Takes All'' (WTA) pipeline to estimate the disparity, offering a better solution.

To summarize, our contributions are as follows.
\begin{enumerate}
	\setlength\itemsep{0em}
	\vspace{-0.1cm}
	\item We propose a novel  deep recurrent model for better handling  stereo disparity estimation tasks. This model generates increasingly  consistent disparities for both views by effectively learning and exploiting online left-right consistency.  To the best of our knowledge, this is the first end-to-end framework unifying consistency check and disparity estimation. 
	\vspace{-0.1cm} 
	\item A soft attention mechanism is introduced to guide the network to automatically focus more on the  unreliable regions  learned by the online left-right comparison during the estimation.
	\vspace{-0.1cm} 
	\item We perform extensive experiments on the KITTI 2015 \cite{Menze2015CVPR, Menze2015ISA}, Scene Flow \cite{mayer2016large} and Middlebury \cite{scharstein2002middlebury} benchmarks to validate the effectiveness of our proposed model, and show that it can achieve state-of-the-art results on these benchmarks.
\end{enumerate}
\section{Related Work}


Traditional stereo matching methods usually utilize the low-level features of image patches around the pixel to measure the dissimilarity. Local descriptors such as absolute difference (AD), sum of squared difference (SSD), census transform \cite{humenberger2010census}, or their combination (AD-CENSUS) \cite{mei2011building} are often employed. For cost aggregation and disparity optimization, some   global methods   treat disparity selection as a multi-label learning problem and optimize a  corresponding 2D graph partitioning problem  by  graph cut \cite{bleyer2007graph} or belief propagation \cite{felzenszwalb2006efficient, sun2003stereo, yang2010constant}. Semi-global methods \cite{hirschmuller2008stereo}  approximately solve  the NP-hard 2D graph partitioning by factorizing it into independent scan-lines and leveraging dynamic programming to aggregate the matching cost.

Deep learning  has  been used in stereo matching recently~\cite{zagoruyko2015learning,zbontar2015computing,luo2016efficient,chen2015deep,shaked2016improved,taniai2016continuous}. Zbontar \etal \cite{zbontar2015computing} trained a siamese network to extract patch features   and then compared  patches accordingly. Luo \etal \cite{luo2016efficient} proposed a faster network model which  computes local matching costs as performing  multi-label classification over disparities.  Chen \etal
\cite{chen2015deep} presented a multi-scale embedding model to obtain faithful local matching costs. Guney \etal \cite{Guney2015CVPR}  introduced object-category specific disparity proposals to better estimate the disparity.  Shaked \etal \cite{shaked2016improved} proposed a constant highway network to learn both 3D cost volume and  matching confidence. Taniai \etal \cite{taniai2016continuous} over-parameterized each pixel with a local disparity plane and constructed an MRF to estimate the disparity map. Combining the matching cost from \cite{zbontar2015computing} and their proposed local expansion move algorithm also achieves  good performance on the Middlebury stereo evaluation \cite{scharstein2002middlebury}.

End-to-end deep learning methods have also been proposed to take the raw images as input and output the final disparity results with deep neural networks without offline post-processing.  Mayer \etal \cite{mayer2016large} leveraged a large dataset to train a convolutional neural network (CNN) to estimate the disparity directly from the image appearances. In ~\cite{mayer2016large}, the cost volume was constructed with the extracted features and the disparity was learned with 3D convolution. GC-NET \cite{kendall2017end} combines the geometric and contextual information with 3D convolution to learn the disparity.
\begin{figure*}
	\begin{minipage}[htbp]{11.1cm}
			\vspace{-0cm}
			\centering
		\includegraphics[width=11cm, height=5cm]{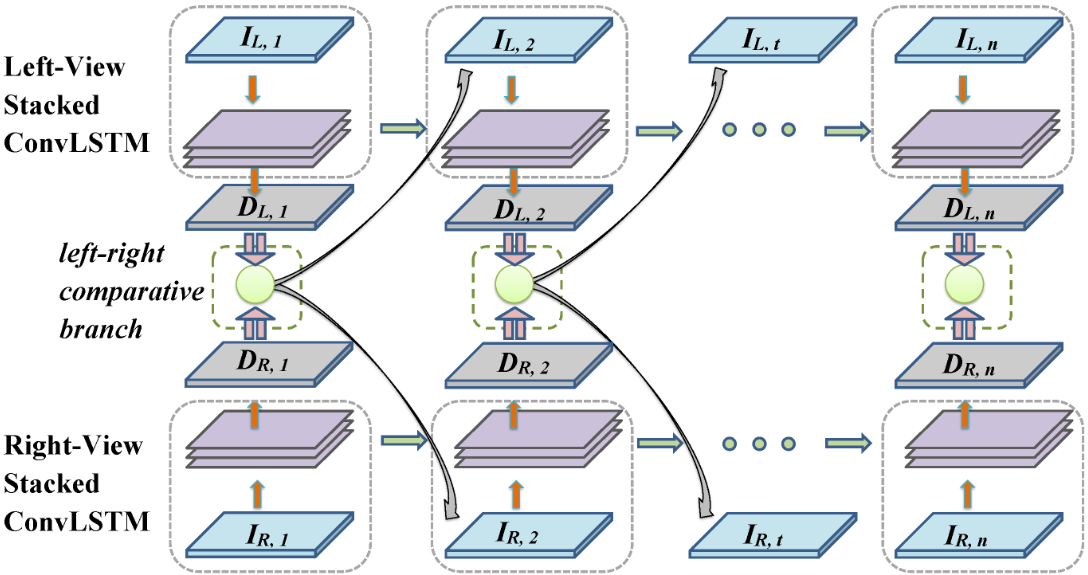}
			\vspace{-0.3cm}
		\caption{\small An illustration of the LRCR model. At each step, the Stacked ConvLSTMs of both views take as  input the corresponding matching cost volume and the error map of that view obtained at the last step. Then, the generated disparity maps are compared to produce the error maps of both views to be fed into the model at the next step, which serve as the soft attention guidance to allow the model to selectively improve the regions.}
		\label{fig:archi}
	
			\vspace{-0.5cm}
		\end{minipage}
		\hspace{0.1cm}
	\begin{minipage}[htbp]{6cm}
			\centering
		\includegraphics[width=6.5cm, height=3cm]{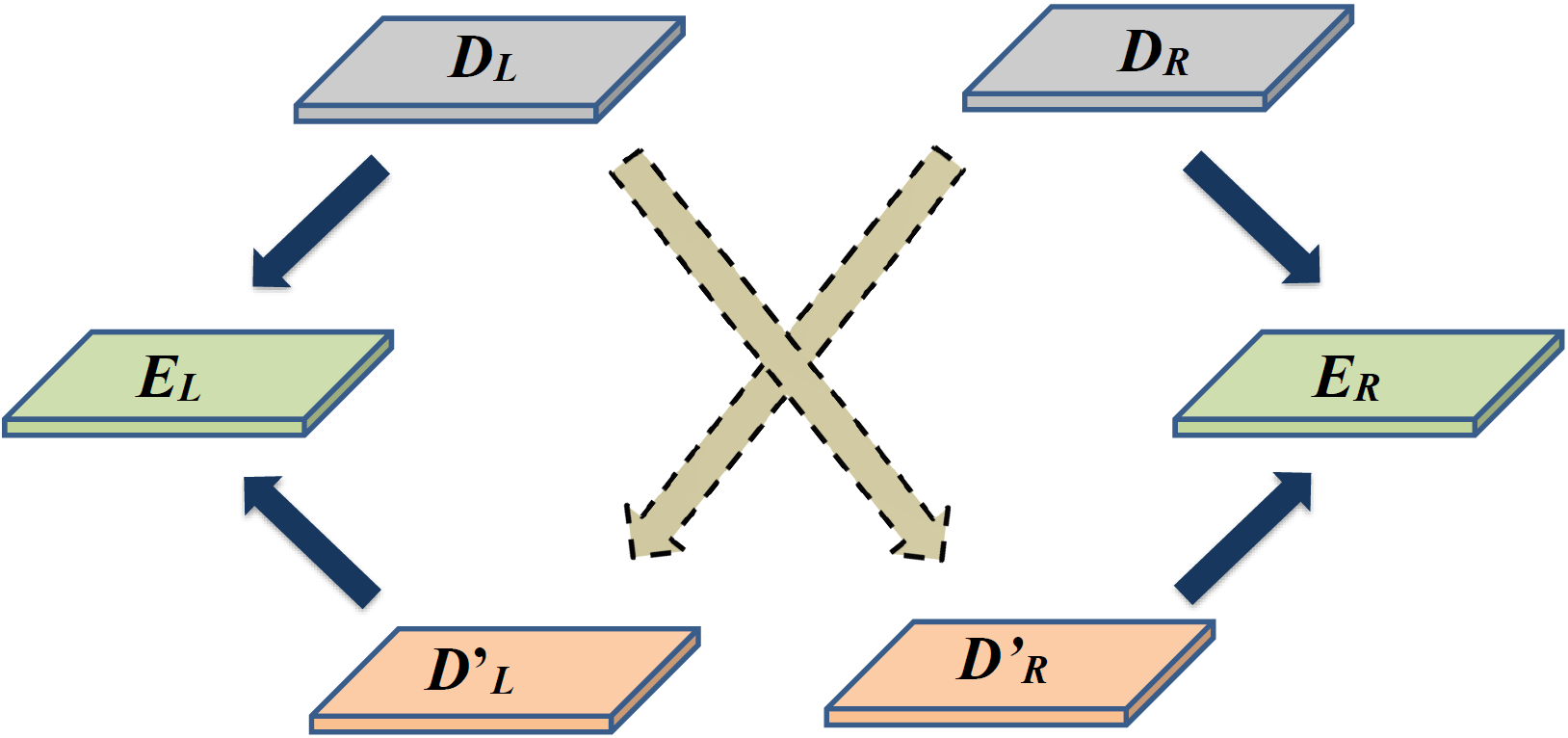}
		\caption{ \small An illustration of the left-right comparative branch. The predicted disparity maps (\emph{i.e.}, $D_{L}$ and $D_{R}$) are first converted to the opposite coordinates to obtain the induced disparity maps (\emph{i.e.}, $D_{L}^{\prime}$ and $D_{R}^{\prime}$). Then $D_{L}$ and $D_{L}^{\prime}$ are fed into a simple network with several convolutional layers to learn the error map $E_{L}$ for the left view, which are the comparison outputs. The same process also applies for the right view.}
		\label{fig:branch}
		\vspace{-0.5cm}
	\end{minipage}

\end{figure*}

Another type of work focuses on getting more reliable  measurements of disparity. A review \cite{hu2012quantitative} has summarized and compared an exhaustive list of hand-crafted confidence measurements for disparity maps.  CNN has also been implemented to improve the accuracy of a confidence prediction by exploiting the local information~\cite{spyropoulos2014learning,haeusler2013ensemble,poggi2017learning,seki2016patch,park2015leveraging,poggi2017quantitative,poggi2016learning}. Haeusler \etal \cite{haeusler2013ensemble} combined different confidence measurements with a random forest classifier to measure the reliability of disparity maps. Seki and Pollefeys \cite{seki2016patch} estimated the confidence map by extracting patches from both the left and right disparity maps with a CNN. Poggi and Mattoccia \cite{poggi2017learning,poggi2016learning} predicted the confidence from only one disparity map within a deep network.

The most closely related work  to ours is \cite{gidaris2016detect},  which also considers  incorporating error localization and improvement.  \cite{seki2016patch} is also related in the sense of detecting left-right mismatched regions as potential erroneous  regions.  However, these methods are only for refinement  and require external disparity results as inputs. By comparison, our method is a unified model integrating disparity generation and left-right comparison without requiring the initial disparity. Our proposed model can generate    disparity maps with increasing left-right  consistency, by learning to focus  on erroneously predicted regions  for disparity enhancement. \cite{godard2017unsupervised} and~\cite{kuznietsov2017semi}  consider left-right consistency in the monocular depth estimation under unsupervised and semi-supervised settings, respectively.

\section{Disparity Estimation with LRCR}

This paper focuses on computing the disparity map for  given rectified stereo pair images. Given  left and  right input images, we aim to produce the corresponding disparity results. The proposed left-right comparative recurrent (LRCR) model takes the matching cost volume of each pixel at all possible disparities as input. It outputs the disparity map containing the disparity values of all the pixels. The matching cost volume is a 3D tensor of size $H \times W \times d_{\max}$, where $H$, $W$, and $d_{\max}$ are the height, the width of the original image, and the maximal possible disparity, respectively. A pixel in the cost volume shows the matching cost between a patch centered around $\mathbf{p} = (\emph{x, y})$ in the left image and a patch centered around $\mathbf{pd} = (\emph{x}-\emph{d}, \emph{y})$,  for every pixel $\mathbf{p}$ in the left image and every possible disparity $\emph{d}$. 

We deploy the constant highway networks \cite{shaked2016improved} to produce the matching cost volume which serves as the input to our proposed LRCR model for disparity estimation. The constant highway network is trained on pairs of small image patches whose true disparity is known. It adopts  a siamese architecture, where each branch processes the left or right image patches individually  and generates its own description vector. Each branch contains several  highway residual blocks with shared weights \cite{shaked2016improved}. Subsequently, two pathways are exploited to compare the two patches and produce a matching cost. In the first pathway, the two feature vectors are concatenated into a single one, and then passed through several fully-connected layers to obtain a binary decision trained via a cross-entropy loss. The second pathway employs a hinge loss to the dot product between the two feature vectors.  Both pathways provide the matching cost volumes that can be taken as input by the later LRCR model for the disparity estimation. During inference, the fully-connected layers are cast into $1 \times 1$ convolutional layers. Taking the whole images as inputs,   the constant highway network outputs the matching cost volumes with the same spatial dimensions in one feed-forward pass efficiently.

\subsection{LRCR Model}
Based on the computed matching cost volume, conventional stereo matching pipelines apply several regularization techniques to incorporate information from  neighboring pixels to obtain smoothed matching costs. Then a simple ``Winner Takes All'' (WTA) rule is implemented to determine the disparity for each pixel. However, such pipelines are still highly hand-crafted and merely uses shallow functions to pool the contextual information from neighboring pixels.

In contrast, we propose a deep model, called as LRCR,  which automatically learns the local contextual information  among the neighborhood for each pixel based on the obtained matching cost volume. Specifically, LRCR  simultaneously improves the disparity results of both views by learning to evaluate the left-right consistency. Formulated as an RNN-style model, the LRCR model can receive the past predicted disparity results and selectively improve them by learning  information about the potential left-right mismatched regions. The progressive improvements facilitate predicting disparity maps with  increasing left-right consistency as desired.

Fig.~\ref{fig:archi} shows the architecture of the proposed LRCR model, which contains two parallel stacked convolutional LSTM networks. The left network takes the cost volume matching left image to right as input and generates a disparity map for the left view. Similarly, the right network generates the disparity map for the right view. At each recurrent step, the two stacked convolutional LSTMs process the two input cost volumes and generate the corresponding disparity maps individually. The two generated disparity maps are then converted to the opposite coordinates \cite{seki2016patch} for comparison with each other. Formally, let $D_{\mathrm{left}, t}$  and $D_{\mathrm{right}, t}$ denote the disparity maps derived from the left-view and  right-view matching cost volumes at the $t^{th}$ step, respectively.   $D_{\mathrm{right}, t}$  can be converted to the left-view coordinates, providing a right-view induced   disparity map $D_{\mathrm{left}, t}^{\prime}$.  Then,  a pixel-wise comparison can be conveniently performed between $D_{\mathrm{left}, t}$ and $D_{\mathrm{left}, t}^{\prime}$ with the information from both views. For doing the comparison,  several convolutional layers and pooling layers are added on the top of  $D_{\mathrm{left, t}}$ and $D_{\mathrm{left, t}}^{\prime}$, producing  the error map of the left-view disparity. Taking such an error map, which serves as a soft attention map (together with the left-view matching cost volume), as  input at the next step, the LRCR model suffices  to selectively focus more on the left-right mismatched  regions at the next step. 



Learning to generate error maps is not trivial. An alternative is to impose the supervision by adopting the difference between the predicted and corresponding ground-truth disparity maps as the regression target. To enhance robustness and solve the aforementioned issues,  we propose  not to  impose  explicit supervision on the error map generation, in order to prevent the left-right comparative branch from learning the simple element-wise subtraction when comparing the two disparity maps. Instead, allowing the network to automatically learn the regions that need to be focused more at the next step may be beneficial to capturing the underlying local correlation between the disparities in the neighborhood. Moreover, we aim to provide a soft attention map whose values serve as confidence or attention weights for the next step prediction, which can hardly be interpreted as a simple element-wise difference between two disparity maps. It is observed that the learned error maps indeed provide reliable results in detecting the mismatched labels which are mostly caused by repetitive patterns, occluded pixels, and depth discontinuities (see Fig. \ref{fig:fig1}). For the right view, the model performs the exactly same process including the conversion of left-view disparity, right-view error map generation, and using the error map as the attention map in the next step of right-view prediction.

\subsection{ Architecture}
We describe the architecture of LRCR model in details, including the stacked convolutional LSTM and the left-right comparative branch. 
\vspace{-0.3cm}
\subsubsection{Stacked Convolutional LSTM}
Convolutional LSTM (ConvLSTM) networks have an advantage in encoding contextual spatial information and reducing the model redundancy, compared to  conventional fully-connected LSTM networks. The LRCR model contains two parallel stacked ConvLSTM  networks that process the left and  right views respectively. The inputs to each stacked ConvLSTM include the matching cost volume of that view and the corresponding error map  generated at the last step. The matching cost volume is a 3D tensor of size $H \times W \times d_{\max}$ and the error map is a 2D map of size  $H \times W $. They are first concatenated along the disparity dimension to form a tensor of size $H \times W \times (d_{\max}+1)$. ConvLSTM is similar to  usual fully-connected LSTM, except that the former applies  spatial convolution operations  on the 2D map  in several layers  to encode   spatial contexts. The detailed unit-wise computations of the ConvLSTM are shown below:
\begin{equation}
\vspace{-0cm}
\begin{aligned}
i_{t} &=  \sigma\big(W_{xi}*X_{t}+W_{hi}*H_{t-1}+W_{ci}\circ C_{t-1}+b_{i}\big), \\
f_{t} &=  \sigma\big(W_{xf}*X_{t}+W_{hf}*H_{t-1}+W_{cf}\circ C_{t-1}+b_{f}\big), \\
o_{t} &=  \sigma\big(W_{xo}*X_{t}+W_{ho}*H_{t-1}+W_{co}\circ C_{t-1}+b_{o}\big), \\
C_{t} &=  f_{t}\circ C_{t-1} + i_{t}\circ \tanh\big(W_{xc}*X_{t}+W_{hc}*H_{t-1}+b_{c}\big), \\
H_{t} &=  o_{t}\circ \tanh(C_{t}).
\end{aligned}
\label{eq:eq1}
\vspace{-0cm}
\end{equation}
Here $ * $ denotes the spatial convolution operation, and $X_{t}$ is the input tensor concatenated by the matching cost volume and the error map generated at the $t^{th}$ step. $H_{t-1}$ and $C_{t-1}$ are the hidden state and memory cell of the $(t-1)^{th}$ step, respectively. $i_{t}$, $f_{t}$ and $o_{t}$ are the gates of the ConvLSTM. Feeding the hidden states of a ConvLSTM as  input to another ConvLSTM successively  builds a stacked ConvLSTM.

The hidden state tensor of the last ConvLSTM is then passed through simple convolutional layers to obtain the cost tensor of size $H \times W \times d_{\max} $. Taking the negative of each value in the cost tensor results in a score tensor. Applying  softmax normalization to the score tensor leads to a probability tensor that reflects the probability on each available disparity for all  pixels. Finally, a differentiable $\arg\min$ layer \cite{kendall2017end} is used to generate the predicted disparity map by summing all the disparities weighted by their probabilities. Mathematically, the following equation describes how one can obtain the predicted disparity $d^{*}$ given the costs on each available disparity $c_{d}$ via the cost tensor for a certain pixel:
\begin{equation}
\vspace{-0.3cm}
d^{*} = \sum_{d=0}^{d_{\max}}d\times\sigma(-c_{d}).
\label{eq:eq2}
\vspace{-0.2cm}
\end{equation}

\subsubsection{Left-Right Comparative Branch}
Fig.~\ref{fig:branch} shows an illustration of the left-right comparative branch. The left-right comparative branch takes as input the left  and  right disparity  maps (\emph{i.e.}, $D_{\mathrm{left}}$ and $D_{\mathrm{right}}$) generated by the left-view and right-view stacked ConvLSTMs, respectively. This branch first converts both maps to the coordinates of the opposite view (\emph{i.e.},  $D_{\mathrm{left}}^{'}$ and $D_{\mathrm{right}}^{\prime}$). Next, the original disparity map and converted disparity map (\emph{e.g.}, $D_{\mathrm{left}}$ and $D_{\mathrm{left}}^{\prime}$) are concatenated and fed into a simple branch consisting of several simple convolutional layers and a final sigmoid transformation layer to obtain the corresponding error map for that view. This error map is then used as the soft attention map in the next step to guide the focused regions for the model.
 \begin{figure*}
	\captionsetup[subfigure]{labelformat=empty}
	\centering
	\hspace{-0.2cm}
	\subfloat[]{\includegraphics[width=3.9cm,height=1.1cm]{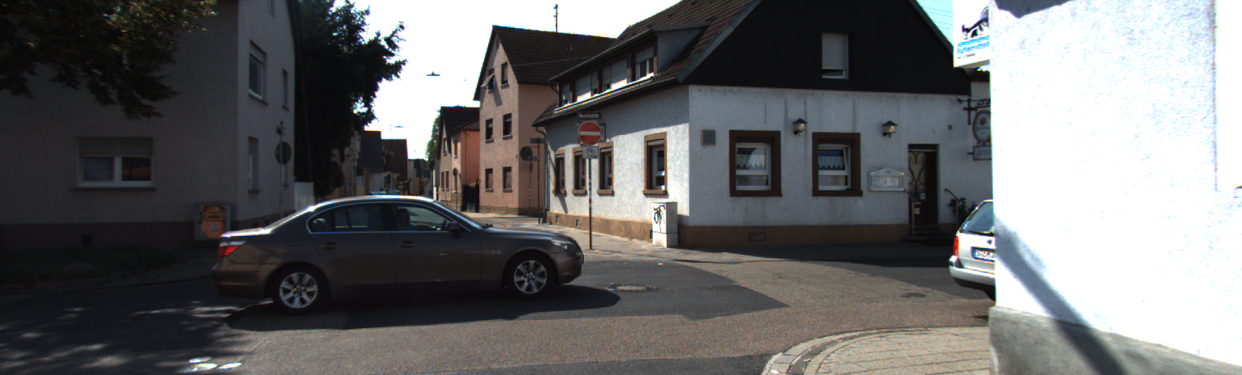}		}
	\hspace{-0.1cm}
	\subfloat[]{\raisebox{0.4cm}{\includegraphics[width=1cm,height=0.6cm]{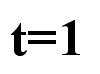}}
	}
	\hspace{-0.2cm}
	\subfloat[]{\includegraphics[width=4cm,height=1.1cm]{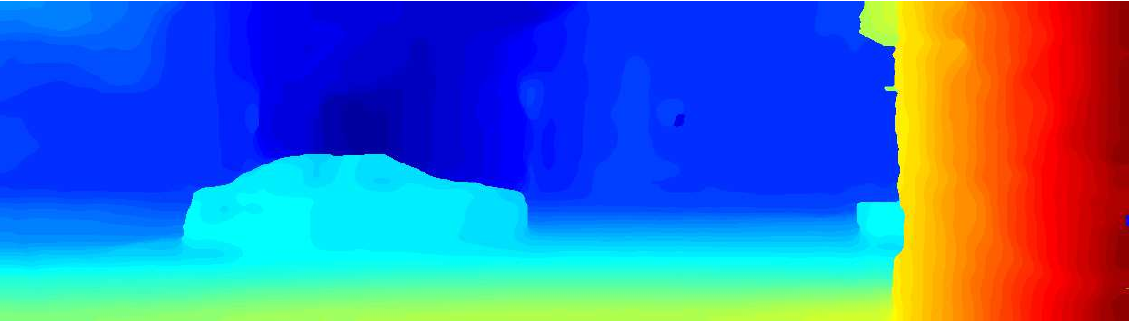}
	}
	\hspace{0cm}
	\subfloat[]{\includegraphics[width=4cm,height=1.1cm]{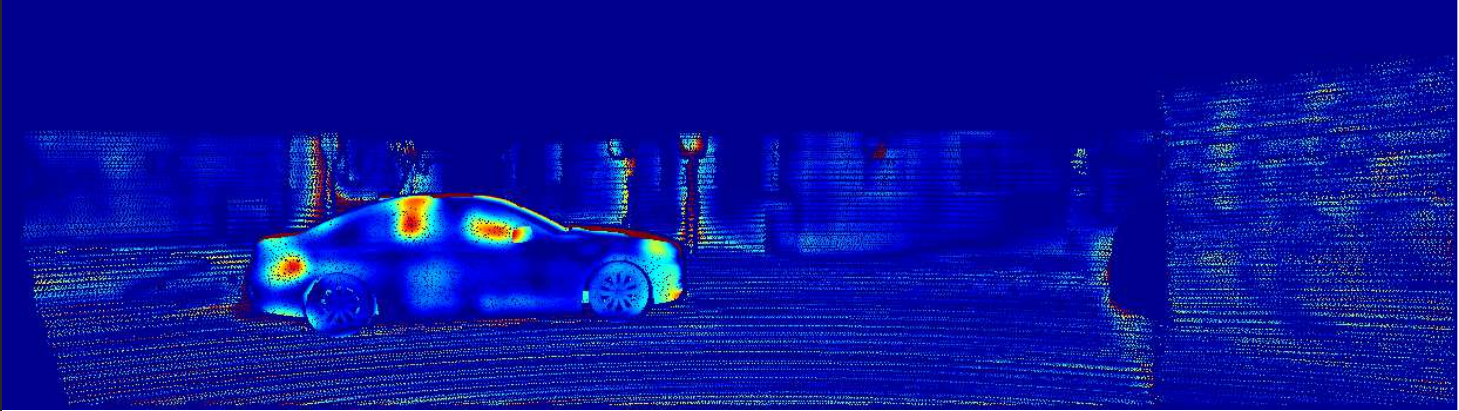}
	}
	\hspace{0cm}
	\subfloat[]{\includegraphics[width=4cm,height=1.1cm]{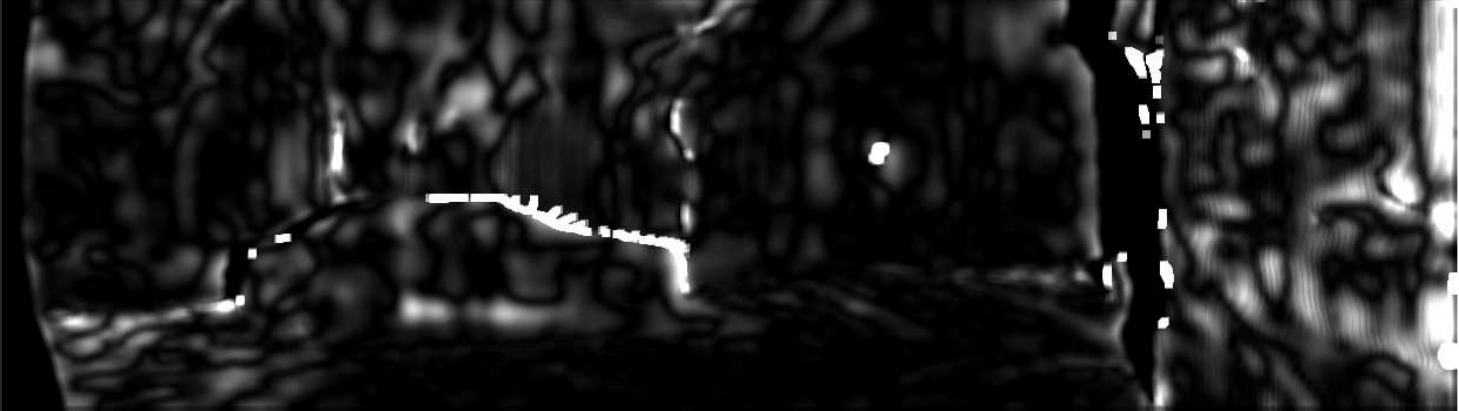}
	}
	\\
	\vspace{-0.7cm}
	\hspace{3.8cm}
	\hspace{-0.1cm}
	\subfloat[]{\raisebox{0.4cm}{\includegraphics[width=1cm,height=0.6cm]{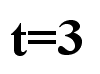}}
	}
	\hspace{-0.2cm}
	\subfloat[]{\includegraphics[width=4cm,height=1.1cm]{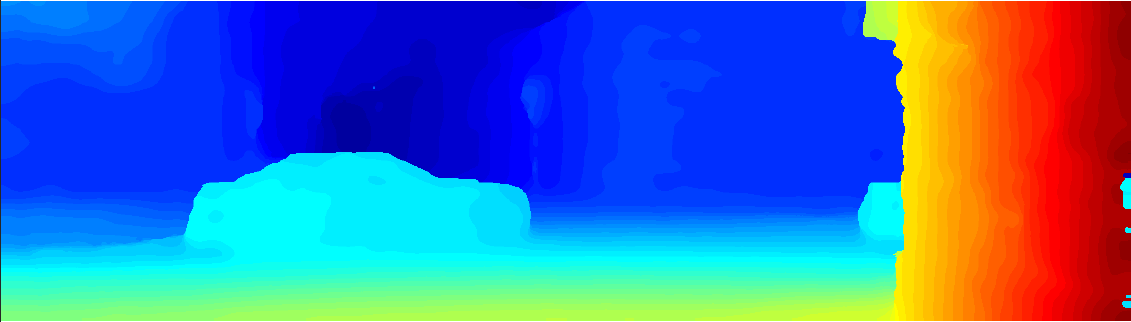}
	}
	\hspace{0cm}
	\subfloat[]{\includegraphics[width=4cm,height=1.1cm]{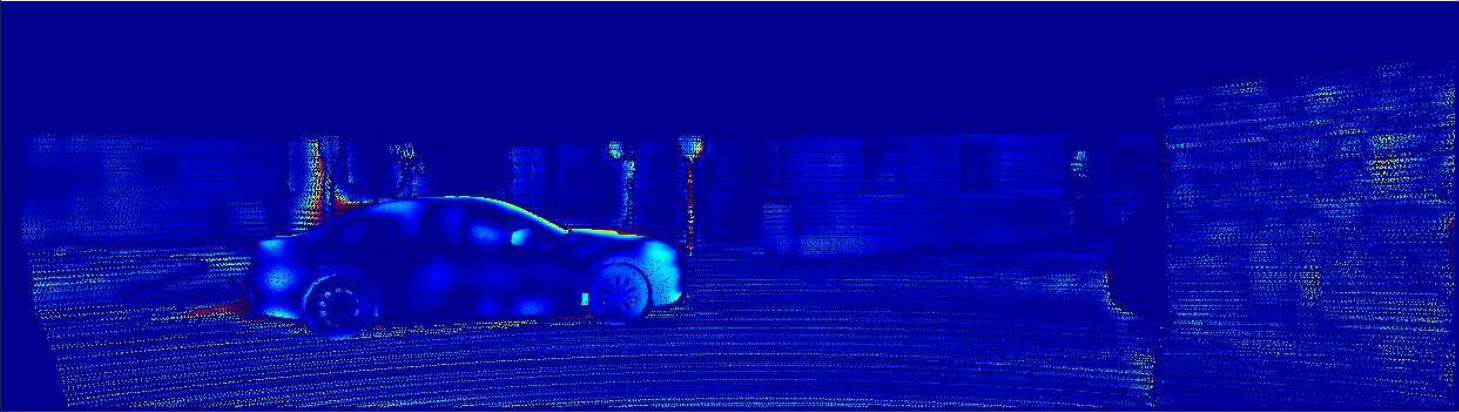}
	}
	\hspace{0cm}
	\subfloat[]{\includegraphics[width=4cm,height=1.1cm]{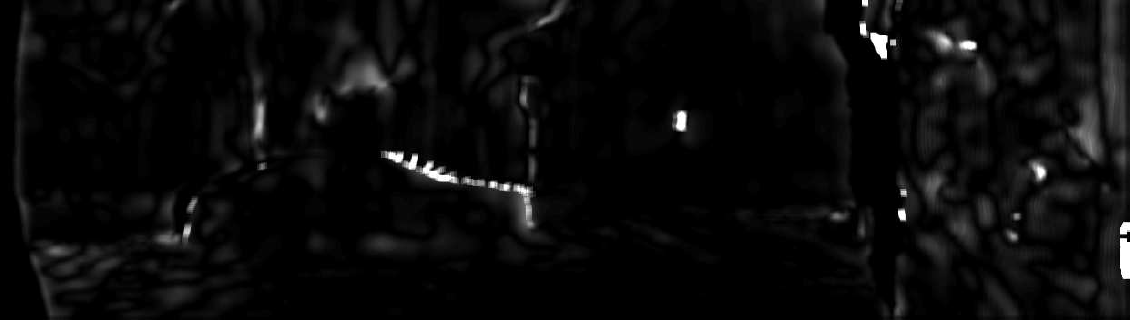}
	}
	\\	
	\vspace{-0.7cm}
	\hspace{3.8cm}
	\hspace{-0.1cm}
	\subfloat[]{\raisebox{0.4cm}{\includegraphics[width=1cm,height=0.6cm]{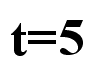}}
	}
	\hspace{-0.2cm}
	\subfloat[]{\includegraphics[width=4cm,height=1.1cm]{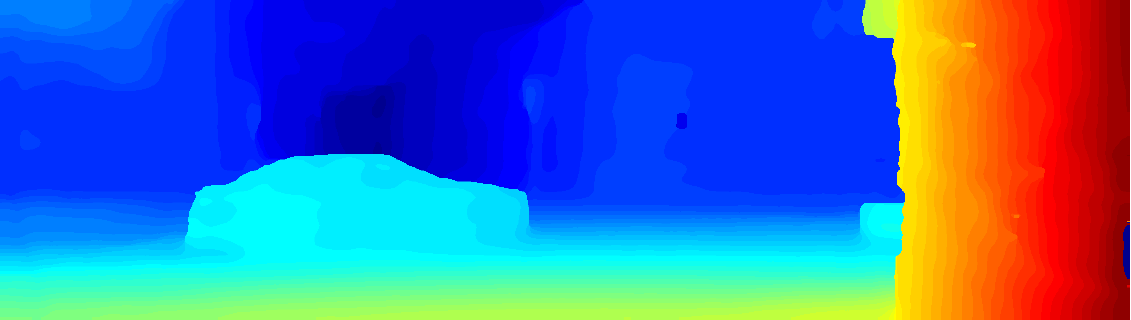}
	}
	\hspace{0cm}
	\subfloat[]{\includegraphics[width=4cm,height=1.1cm]{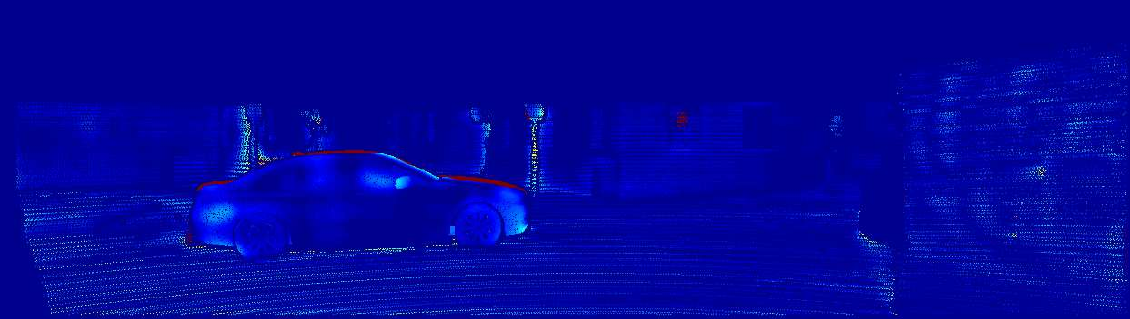}
	}
	\hspace{0cm}
	\subfloat[]{\includegraphics[width=4cm,height=1.1cm]{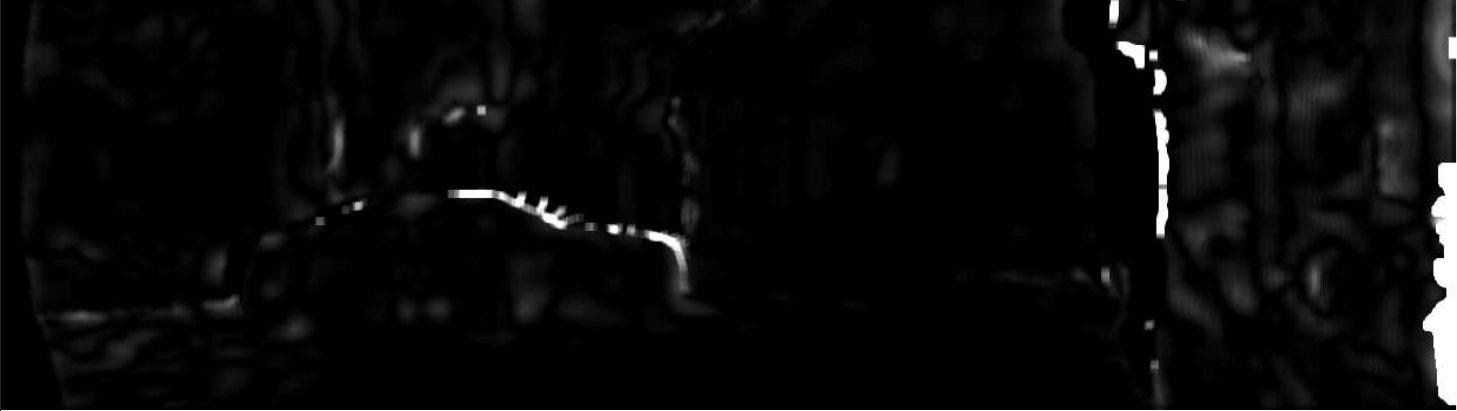}
	}
	\\
	\vspace{-0.6cm}
	\hspace{-0.2cm}
	\subfloat[]{\includegraphics[width=3.9cm,height=1.1cm]{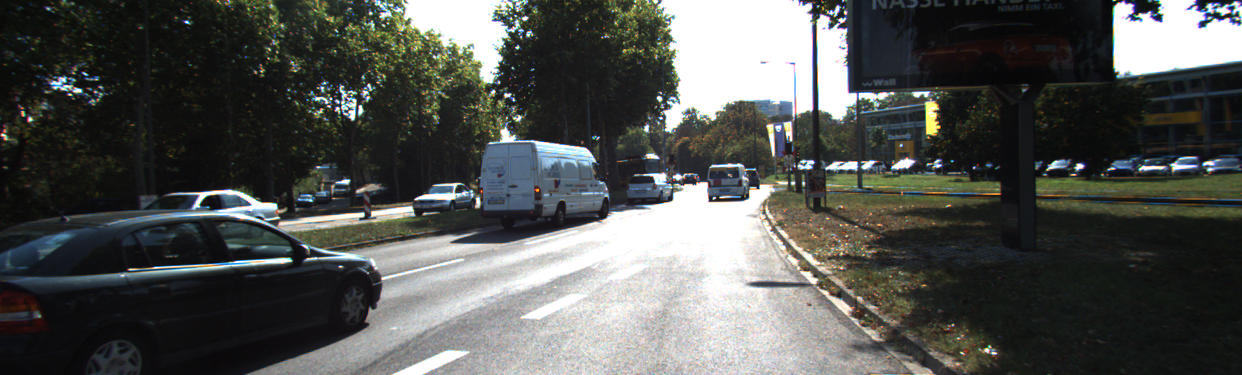}		}
	\hspace{-0.1cm}
	\subfloat[]{\raisebox{0.4cm}{\includegraphics[width=1cm,height=0.6cm]{./figures/t1.png}}
	}
	\hspace{-0.2cm}
	\subfloat[]{\includegraphics[width=4cm,height=1.1cm]{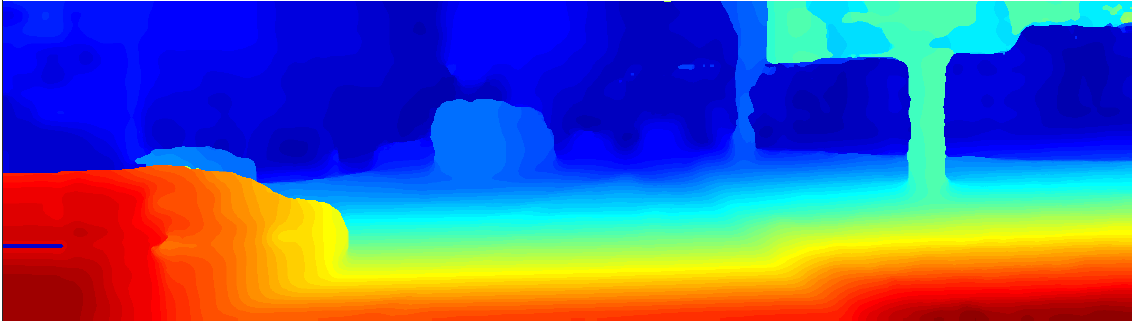}
	}
	\hspace{0cm}
	\subfloat[]{\includegraphics[width=4cm,height=1.1cm]{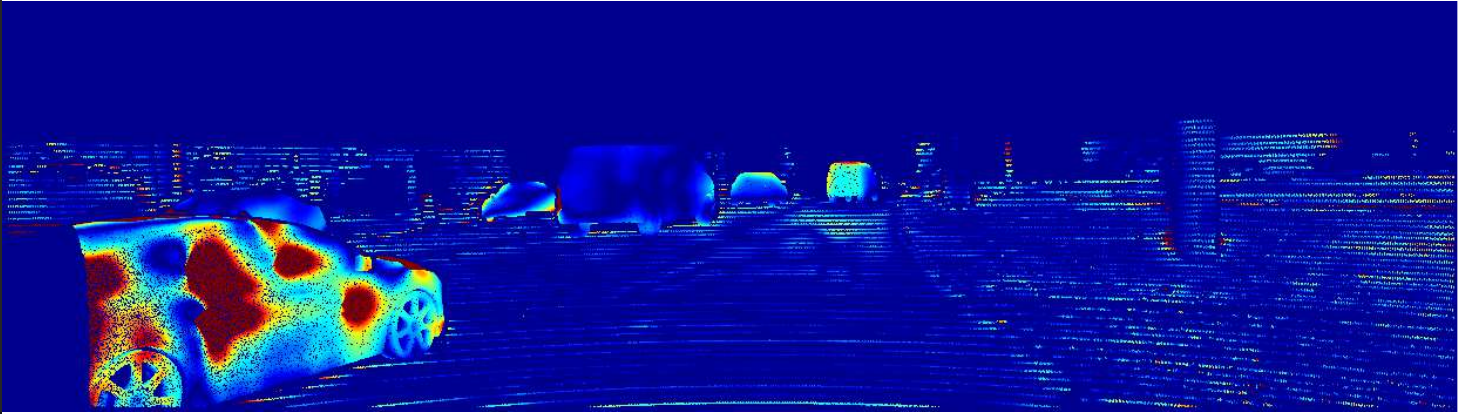}
	}
	\hspace{0cm}
	\subfloat[]{\includegraphics[width=4cm,height=1.1cm]{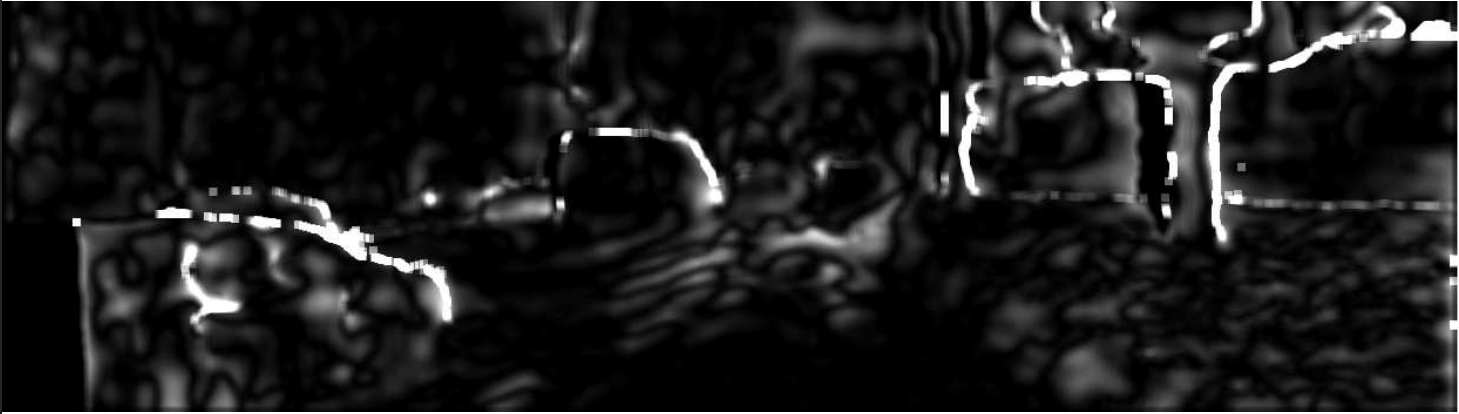}
	}
	\\
	\vspace{-0.7cm}
	\hspace{3.8cm}
	\hspace{-0.1cm}
	\subfloat[]{\raisebox{0.4cm}{\includegraphics[width=1cm,height=0.6cm]{./figures/t3.png}}
	}
	\hspace{-0.2cm}
	\subfloat[]{\includegraphics[width=4cm,height=1.1cm]{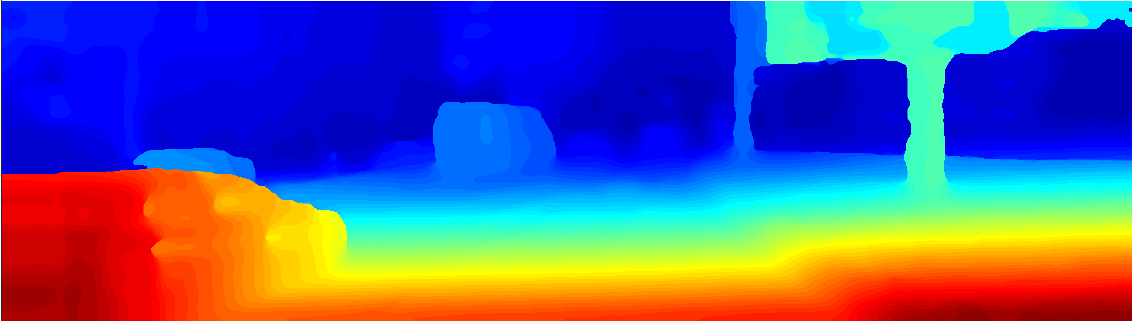}
	}
	\hspace{0cm}
	\subfloat[]{\includegraphics[width=4cm,height=1.1cm]{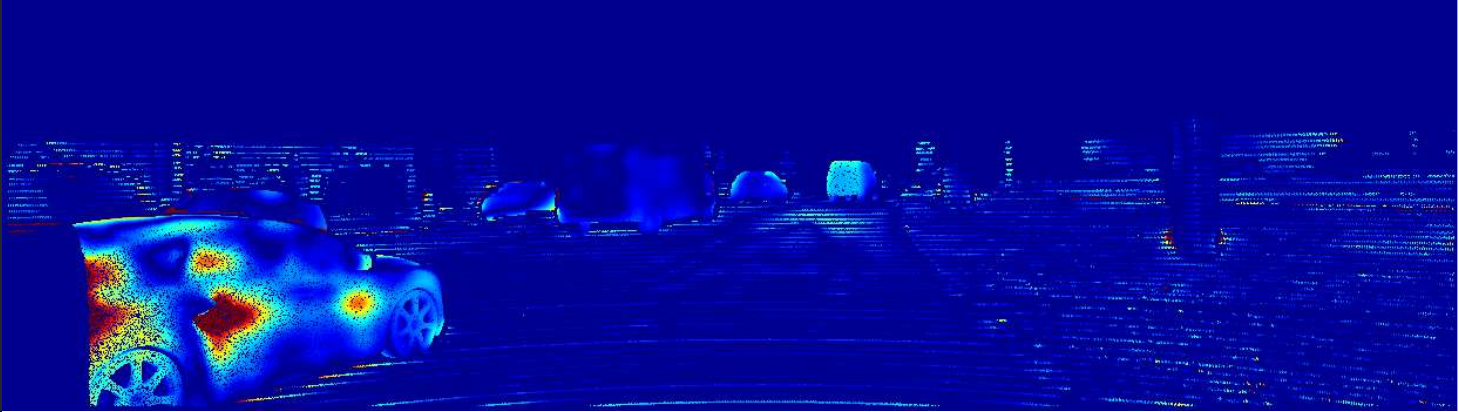}
	}
	\hspace{0cm}
	\subfloat[]{\includegraphics[width=4cm,height=1.1cm]{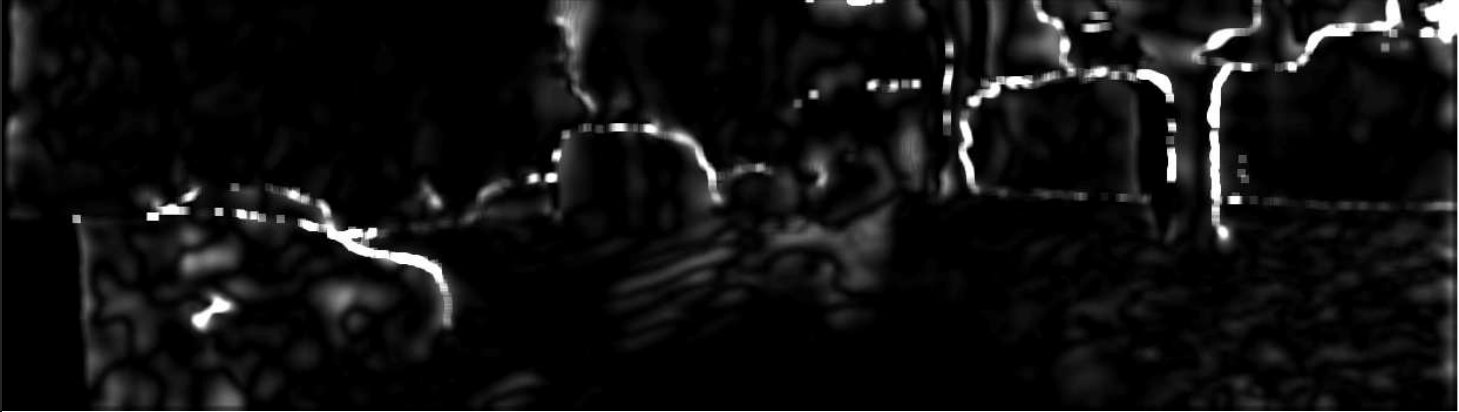}
	}
	\\	
	\vspace{-0.7cm}
	\hspace{3.8cm}
	\hspace{-0.1cm}
	\subfloat[]{\raisebox{0.4cm}{\includegraphics[width=1cm,height=0.6cm]{./figures/t5.png}}
	}
	\hspace{-0.2cm}
	\subfloat[]{\includegraphics[width=4cm,height=1.1cm]{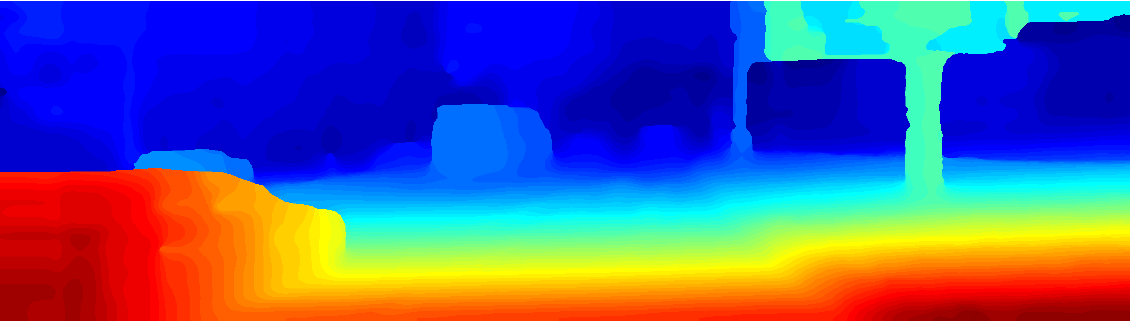}
	}
	\hspace{0cm}
	\subfloat[]{\includegraphics[width=4cm,height=1.1cm]{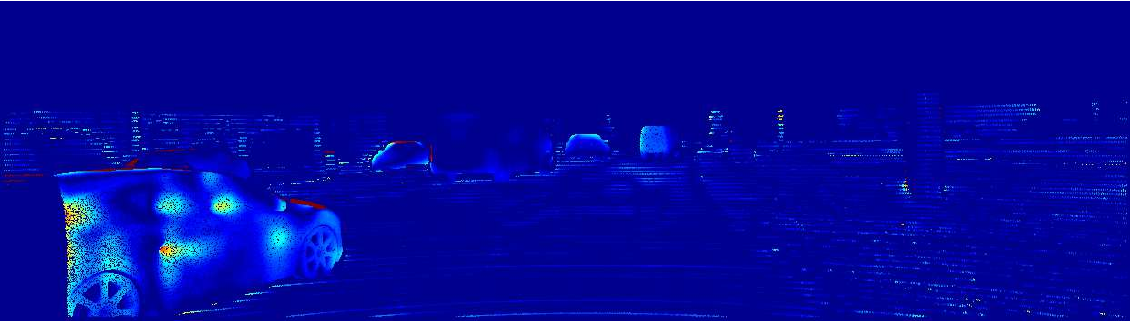}
	}
	\hspace{0cm}
	\subfloat[]{\includegraphics[width=4cm,height=1.1cm]{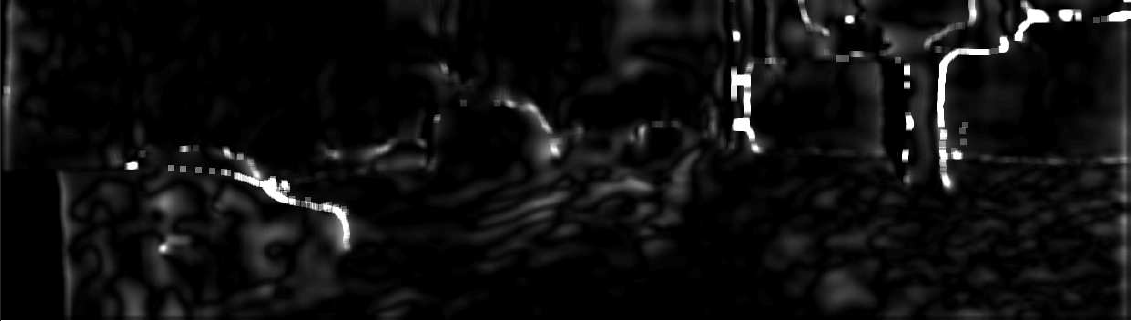}
	}
	\\
	\vspace{-0.6cm}
	\hspace{-0.2cm}
	\subfloat[]{\includegraphics[width=3.9cm,height=1.1cm]{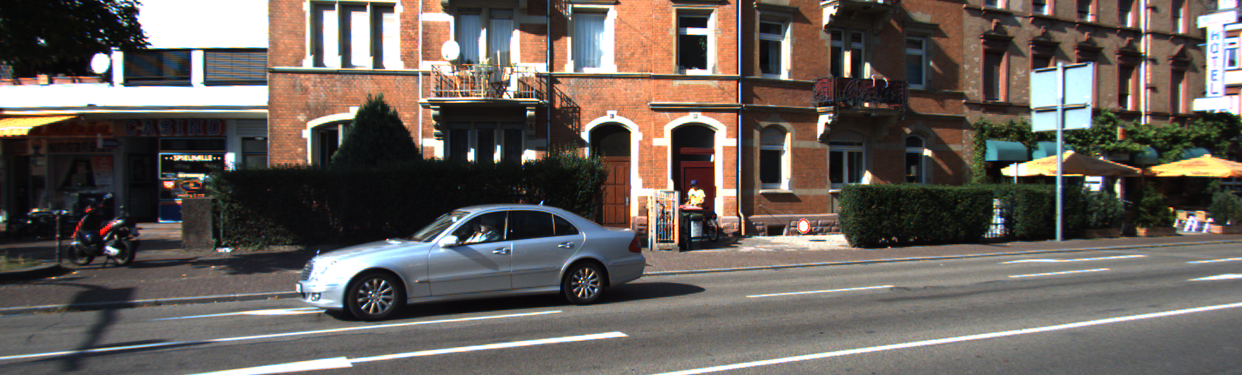}		}
	\hspace{-0.1cm}
	\subfloat[]{\raisebox{0.4cm}{\includegraphics[width=1cm,height=0.6cm]{./figures/t1.png}}
	}
	\hspace{-0.2cm}
	\subfloat[]{\includegraphics[width=4cm,height=1.1cm]{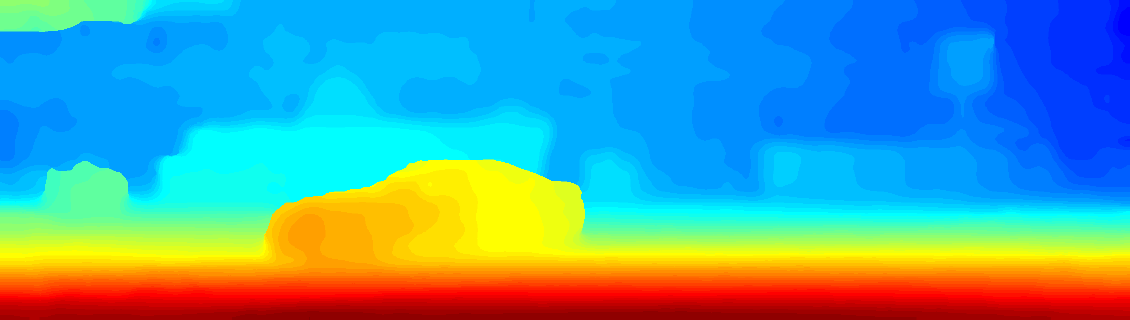}
	}
	\hspace{0cm}
	\subfloat[]{\includegraphics[width=4cm,height=1.1cm]{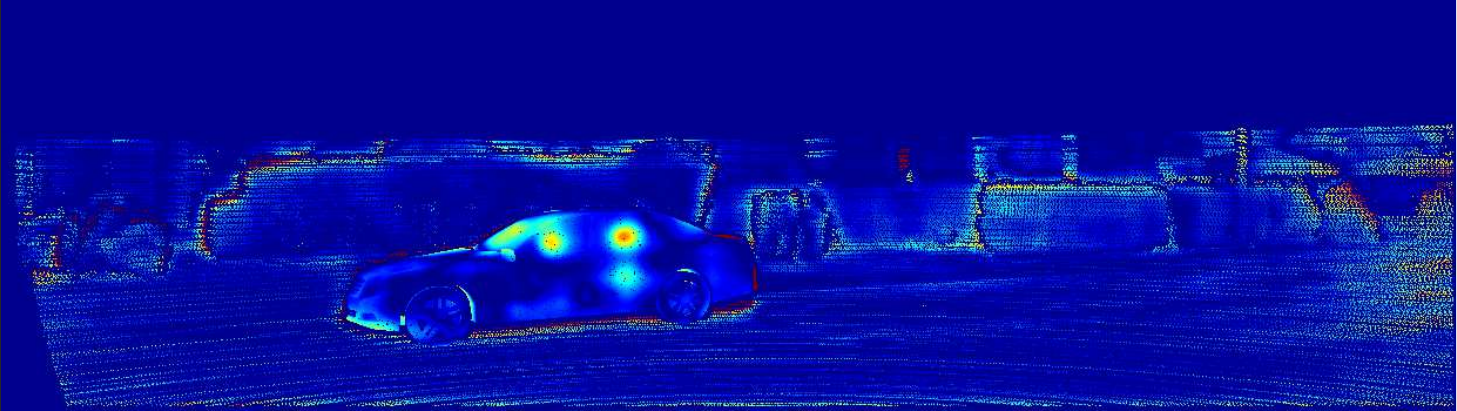}
	}
	\hspace{0cm}
	\subfloat[]{\includegraphics[width=4cm,height=1.1cm]{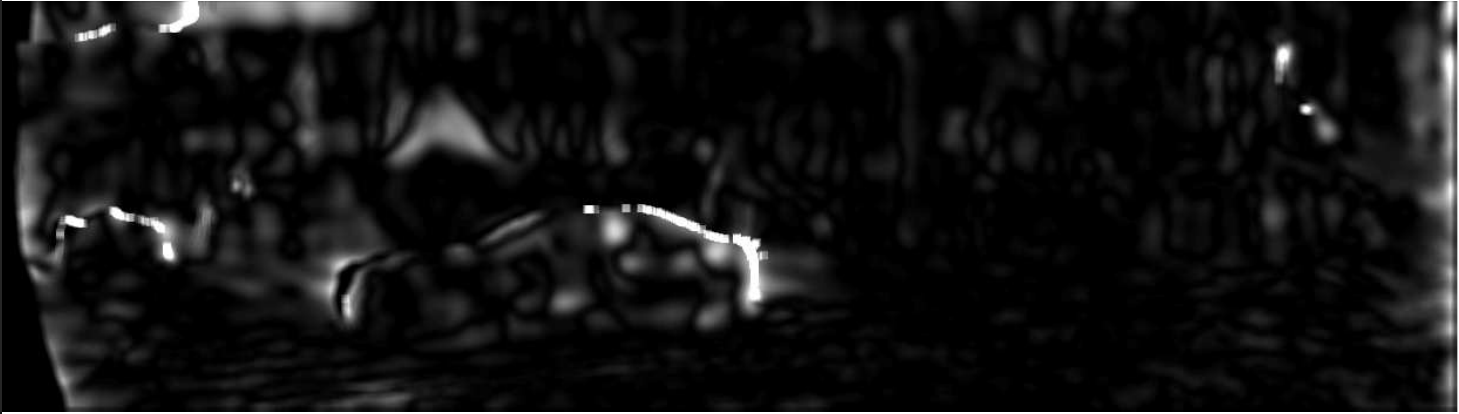}
	}
	\\
	\vspace{-0.7cm}
	\hspace{3.75cm}
	\hspace{-0.1cm}
	\subfloat[]{\raisebox{0.4cm}{\includegraphics[width=1cm,height=0.6cm]{./figures/t3.png}}
	}
	\hspace{-0.2cm}
	\subfloat[]{\includegraphics[width=4cm,height=1.1cm]{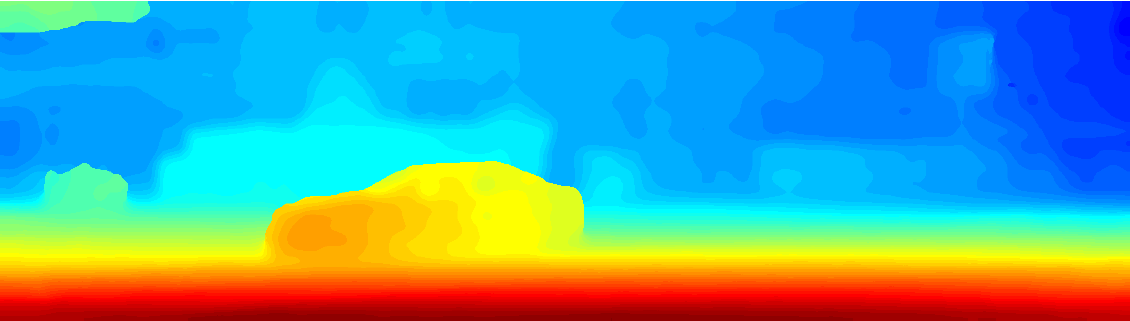}
	}
	\hspace{0.03cm}
	\subfloat[]{\includegraphics[width=4cm,height=1.1cm]{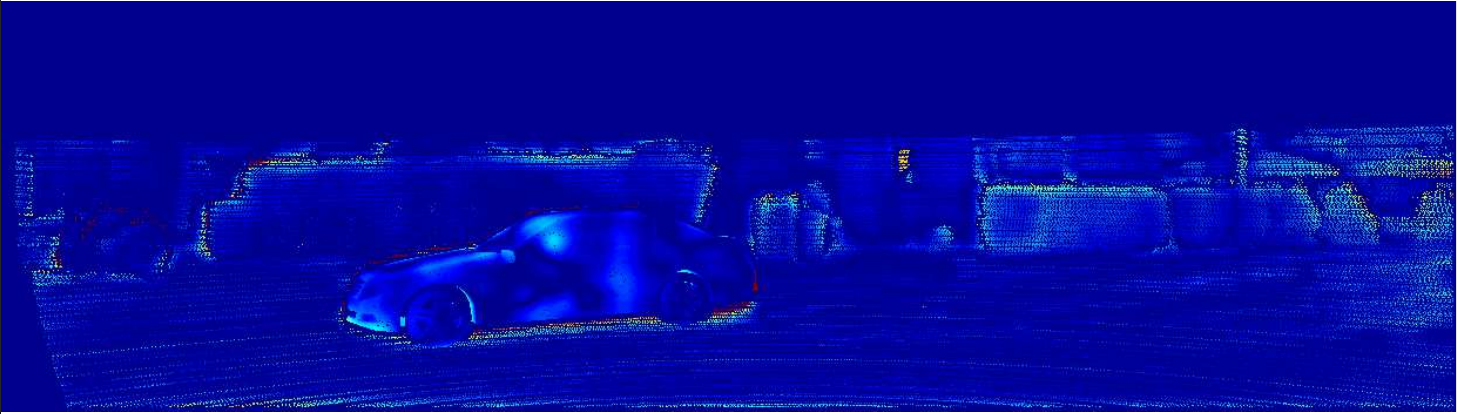}
	}
	\hspace{0cm}
	\subfloat[]{\includegraphics[width=4cm,height=1.1cm]{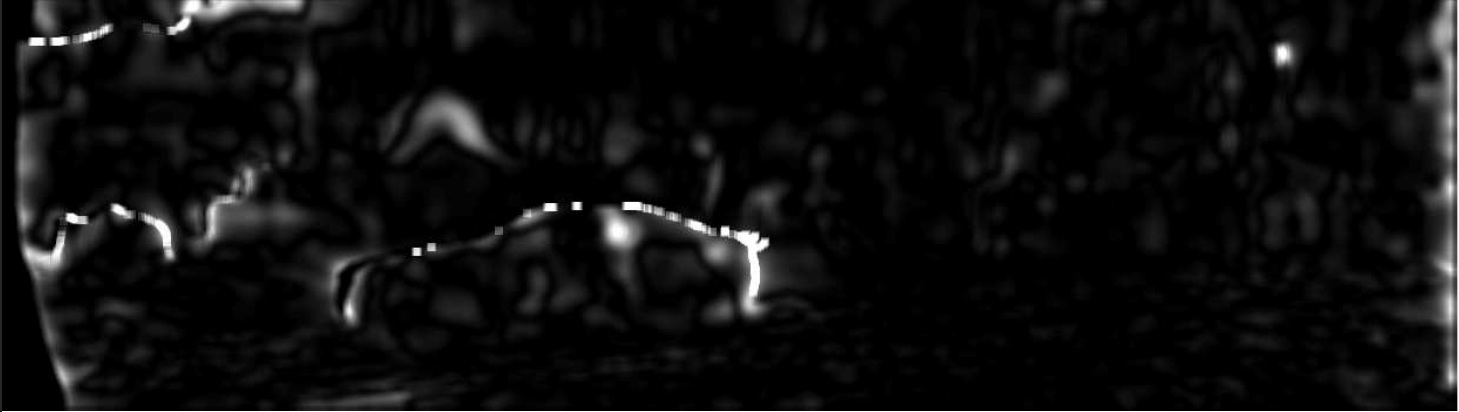}
	}
	\\	
	\vspace{-0.7cm}
	\hspace{3.8cm}
	\hspace{-0.1cm}
	\subfloat[]{\raisebox{0.4cm}{\includegraphics[width=1cm,height=0.6cm]{./figures/t5.png}}
	}
	\hspace{-0.2cm}
	\subfloat[]{\includegraphics[width=4cm,height=1.1cm]{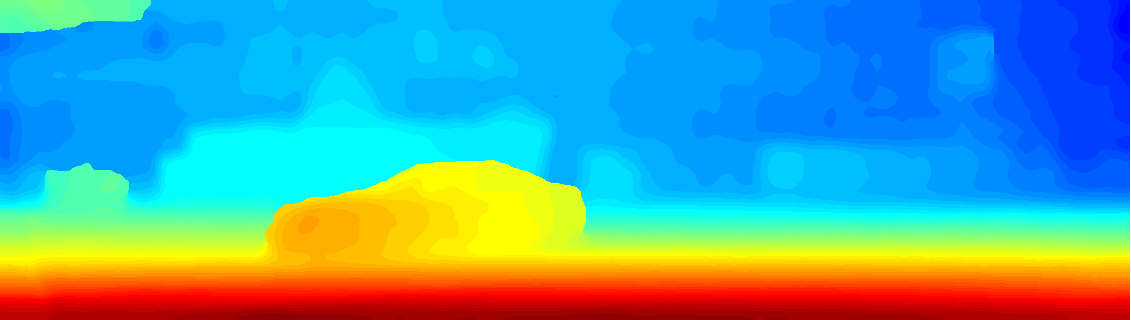}
	}
	\hspace{0cm}
	\subfloat[]{\includegraphics[width=4cm,height=1.1cm]{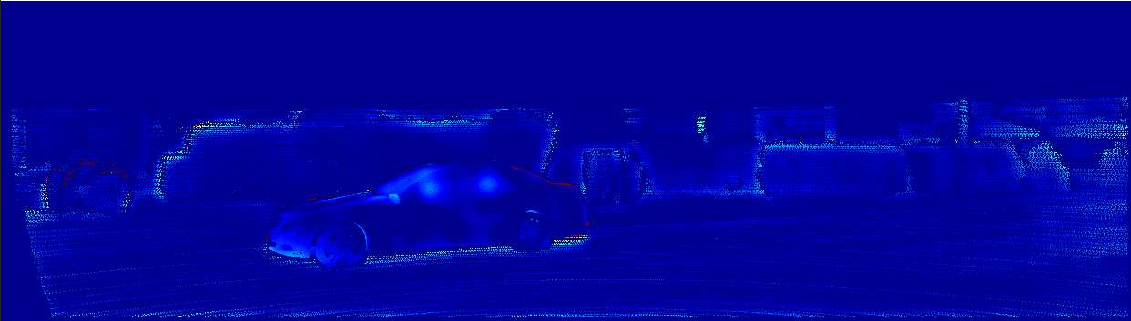}
	}
	\hspace{0cm}
	\subfloat[]{\includegraphics[width=4cm,height=1.1cm]{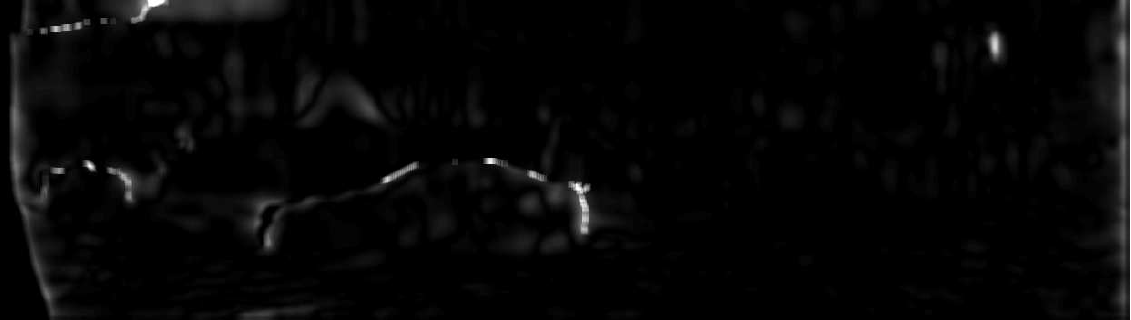}
	}
	\\
	\vspace{-0.8cm}
	\hspace{-0cm}
	\subfloat[]{\includegraphics[width=2.5cm,height=0.8cm]{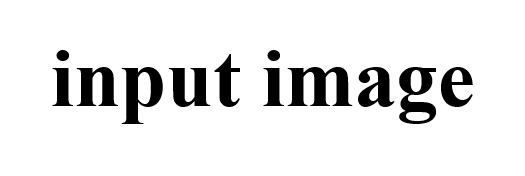}
	}
	\hspace{2.5cm}
	\subfloat[]{\includegraphics[width=2cm,height=0.8cm]{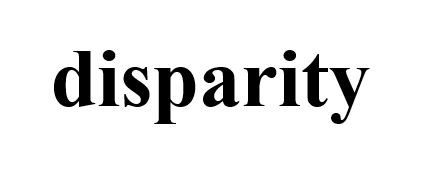}
	}
	\hspace{2.1cm}
	\subfloat[]{\includegraphics[width=2cm,height=0.8cm]{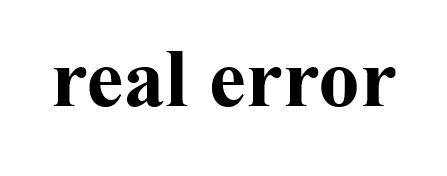}
	}
	\hspace{1.6cm}
	\subfloat[]{\includegraphics[width=2.8cm,height=0.8cm]{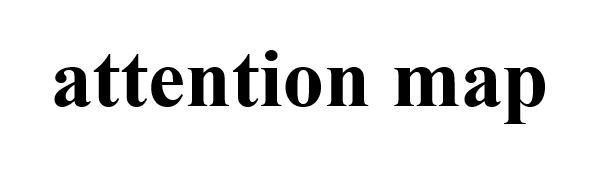}
	}
	\\
	\vspace{-0.8cm}
	\caption{Visualization of input images, predicted disparity maps, real error maps (difference from the ground-truth disparity maps), and the learned attention maps at different recurrent steps. Note that the attention maps of the 5th step are obtained by forwarding the final disparity (obtained at the 5th step) to the left-right comparative branch once more.}
	\label{fig:comp}
	\vspace{-0.6cm}
\end{figure*}

\subsection{Training}
\label{sec:training}
The supervisions of the whole ground-truth disparity maps are imposed on the predicted disparity maps for both left and right views at each recurrent step. The  $\ell_1$ norm of the difference between the ground-truth and the predicted disparity is used as the training loss. The averaged loss over all the labeled pixels is optimized for a particular sample. The loss is mathematically defined in \eqref{eq:eq3}, where $N$ is the number of labeled pixels for the sample, and $d_{n}$ and $d_{n}^{*}$ are the predicted disparity and the ground-truth disparity for the $n^{th}$ pixel, respectively.
\begin{equation}
\vspace{-0.1cm}
L = \frac{1}{N}\sum_{n=1}^{N}\big||d_{n}-d_{n}^{*}\big||_{1}.
\label{eq:eq3}
\vspace{-0.2cm}
\end{equation}
Due to the complexity of the model architecture, we train the whole LRCR model in two stages. We expect the model trained by the first stage to provide reliable initial disparity estimations, while the second stage trains the model to progressively improve the estimations. At the first stage, we  train the stacked ConvLSTMs with only one recurrent step. Indeed, the stacked ConvLSTM performs as a feed-forward network that receives only the original matching cost volumes and predicts disparity by pooling local context via spatial convolutions, similar to \cite{kendall2017end} and \cite{zhong2017self}. In this stage, $H_{t-1}$ and $C_{t-1}$ are set as  constant zero tensors. Therefore, only $W_{xi}$, $W_{xf}$ and $W_{xc}$ are trained in this stage, and the rest weights in \eqref{eq:eq1} will not be trained. This stage results in a non-recurrent version of the LRCR model. In the second stage, the non-recurrent version of the LRCR model trained previously is used as the initialization to the full LRCR model. All the weights in \eqref{eq:eq1} as well as the left-right comparative branch will be trained in this stage.

 \section{Experiments}
 We extensively evaluate our LRCR model  on three largest and   most challenging stereo benchmarks: KITTI 2015, Scene Flow and Middlebury. In Sec.~\ref{sec:experimentKITTI}, we experiment with different variants of our model and provide detailed component analysis on the KITTI 2015 dataset. In Sec.~\ref{sec:experimentSCENEFLOW}, we conduct the experiments on the Scene Flow dataset. In Sec.~\ref{sec:experimentMIDDLEBURY}, we report the results on the Middlebury benchmark and also compare our model against the  state-of-the-arts. In all the experiments, after generating the disparity maps by the LRCR model, sub-pixel enhancement plus median and bilateral filterings is used to refine the results.
 
 \subsection{Experiments on KITTI 2015}
 \label{sec:experimentKITTI}
 \paragraph{Dataset.} KITTI 2015 is a real-world dataset with dynamic street views captured by a driving car. It provides 200 stereo pairs with sparse ground-truth disparities for training and 200 pairs for testing through online leaderboard. We randomly split a validation set with 40 pairs from the training set. When evaluated on the validation set, our model is trained on the training set containing the rest 160 pairs. When doing the evaluation on the testing set, we train the model on the whole training set.

 \paragraph{Implementation details.} The constant highway network~\cite{shaked2016improved} is used to learn the matching cost volumes as input to our LRCR model. It has two versions, \emph{i.e.,} the hybrid one and the fast one. The hybrid version contains two pathways trained by both the cross-entropy loss and the hinge loss in the matching cost learning, while the fast one has only the dot-product pathway trained by the hinge loss. $d_{\mathrm{max}}$ is set to 228. In our LRCR model, each  of the parallel stacked ConvLSTMs contains four ConvLSTM layers. The numbers of  output channels in the four ConvLSTM layers are $d_{\mathrm{max}}$, $2d_{\mathrm{max}}$, $2d_{\mathrm{max}}$, and $d_{\mathrm{max}}$, respectively. All the convolution kernel sizes in the four ConvLSTM layers are $3*3$, and $1*1$ paddings are applied to keep the feature map sizes. After the four ConvLSTM layers, additional three $1*1$ convolutional layers are included and followed by the final differentiable $\arg\min$ layer. The numbers of output channels in the three $1*1$ convolutional layers are $d_{\mathrm{max}}$, $d_{\mathrm{max}}$, and $d_{\mathrm{max}}$, respectively. The left-right comparative branch has two $3*3$ convolutional layers with $1*1$ padding.

 As explained in Sec.~\ref{sec:training}, we first perform the training of the non-recurrent version of our LRCR model. This stage lasts for 15 epochs. The initial learning rate is set to 0.01 and decreased by a factor of 10 every 5 epochs. In the second stage, the well trained non-recurrent version of the LRCR model is used for the weight initialization. We train the recurrent model for 5 recurrent steps. We train 30 epochs in this stage. The initial learning rate is set to 0.01 and decreased by a factor of 10 every 10 epochs. 
\begin{figure*}
	\captionsetup[subfigure]{labelformat=empty}
	\centering
	\hspace{-0.2cm}
	\vspace{-0cm}
	\subfloat[]{\includegraphics[width=5.5cm,height=1.5cm]{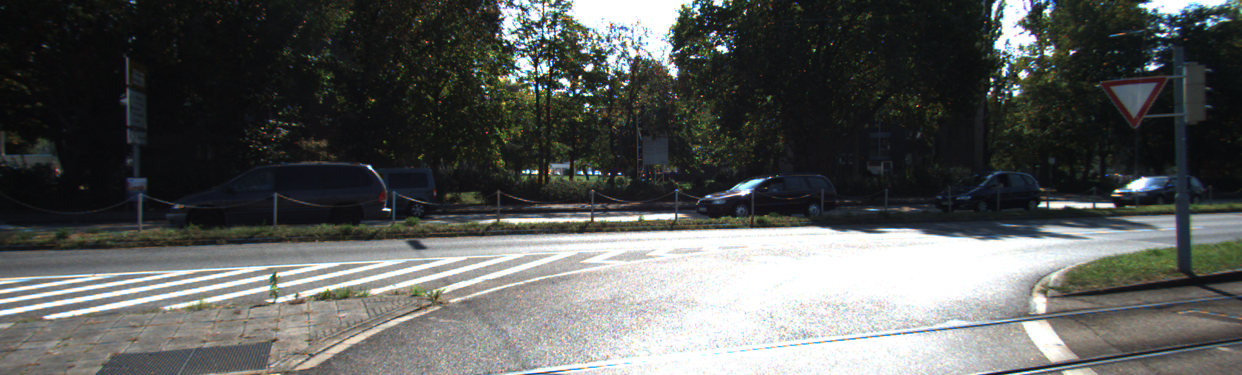}		}
	\hspace{0cm}
	\subfloat[]{\includegraphics[width=5.5cm,height=1.5cm]{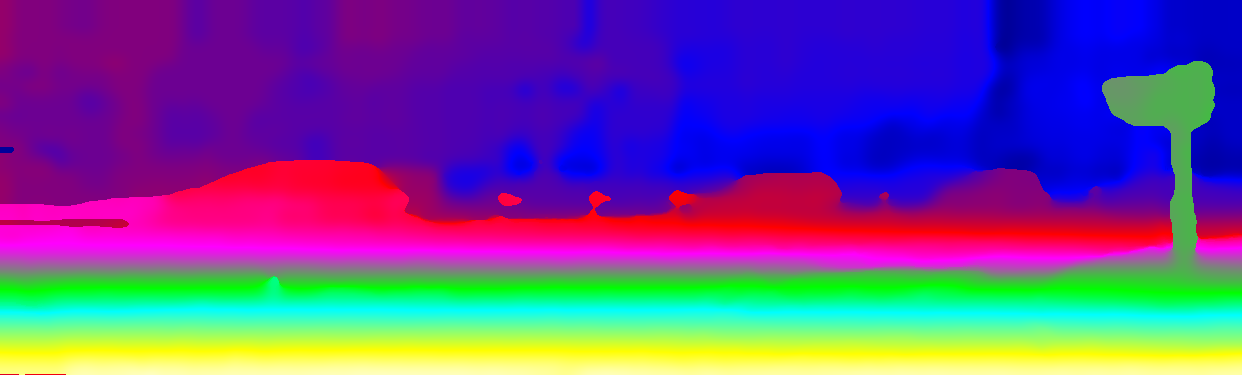}
	}
	\hspace{0cm}
	\subfloat[]{\includegraphics[width=5.5cm,height=1.5cm]{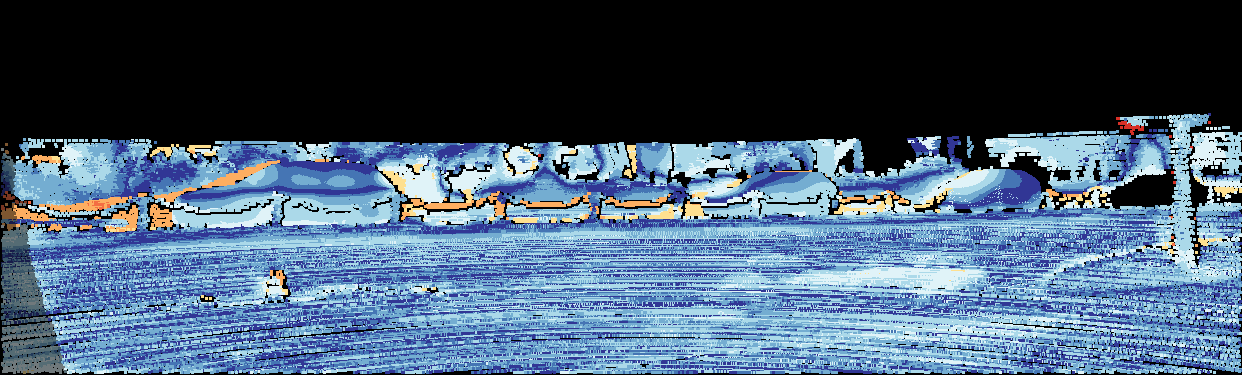}
	}
	\\
	\vspace{-0.7cm}
	\hspace{-0.2cm}
	\subfloat[]{\includegraphics[width=5.5cm,height=1.5cm]{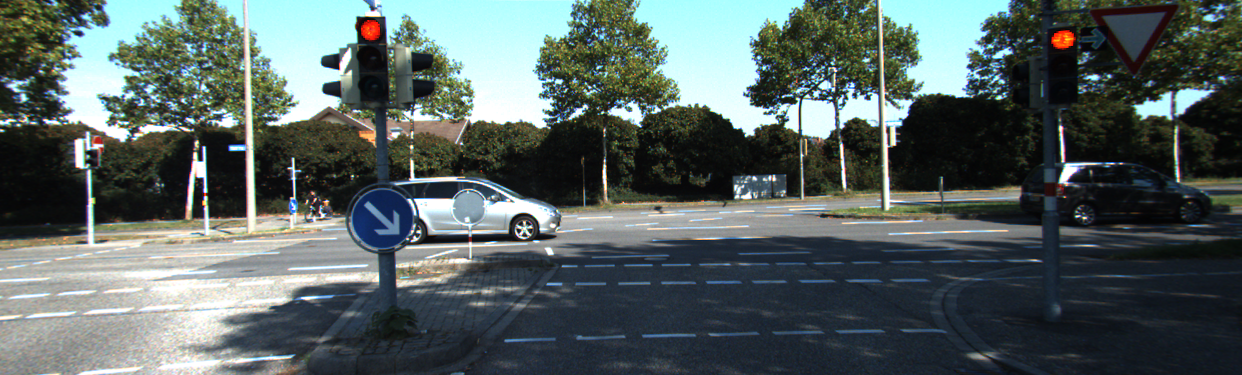}
	}
	\hspace{0cm}
	\subfloat[]{\includegraphics[width=5.5cm,height=1.5cm]{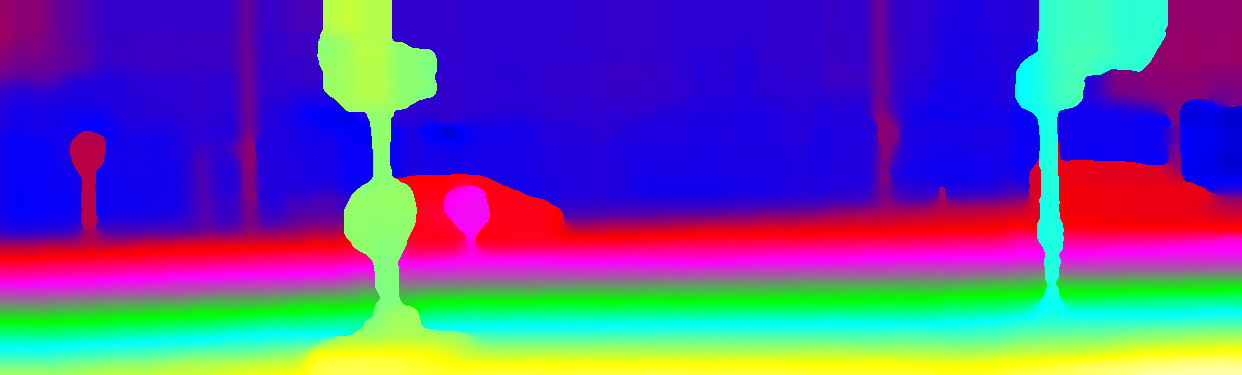}		}
	\hspace{0cm}
	\subfloat[]{\includegraphics[width=5.5cm,height=1.5cm]{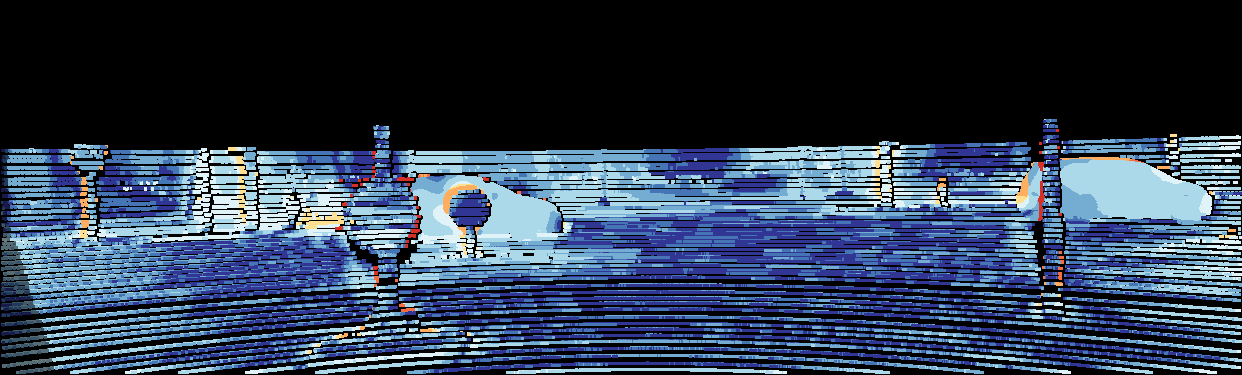}
	}
	\\	
	\vspace{-0.7cm}
	\hspace{-0.2cm}
	\subfloat[]{\includegraphics[width=5.5cm,height=1.5cm]{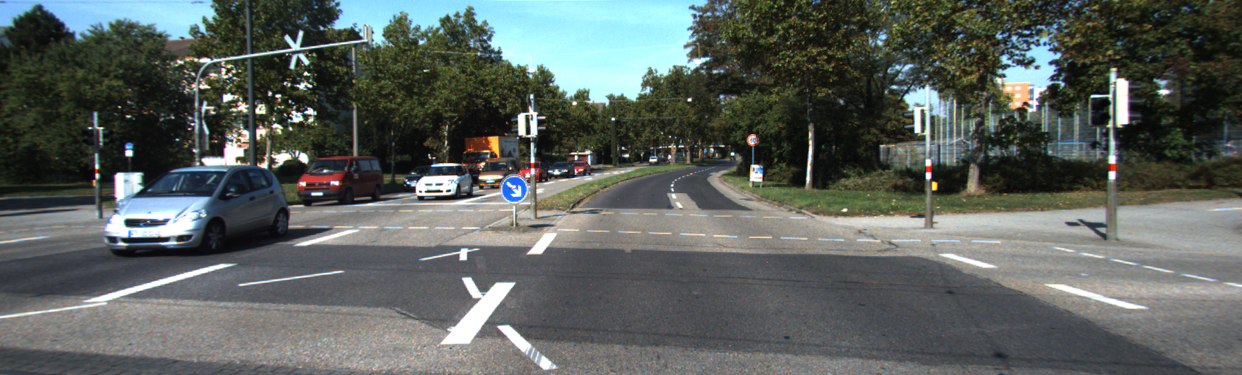}
	}
	\hspace{0cm}
	\subfloat[]{\includegraphics[width=5.5cm,height=1.5cm]{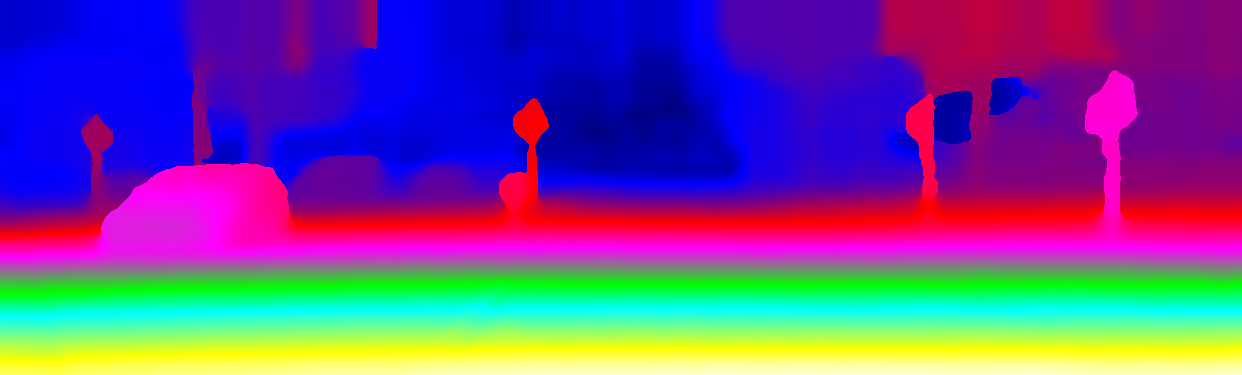}
	}
	\hspace{0cm}
	\subfloat[]{\includegraphics[width=5.5cm,height=1.5cm]{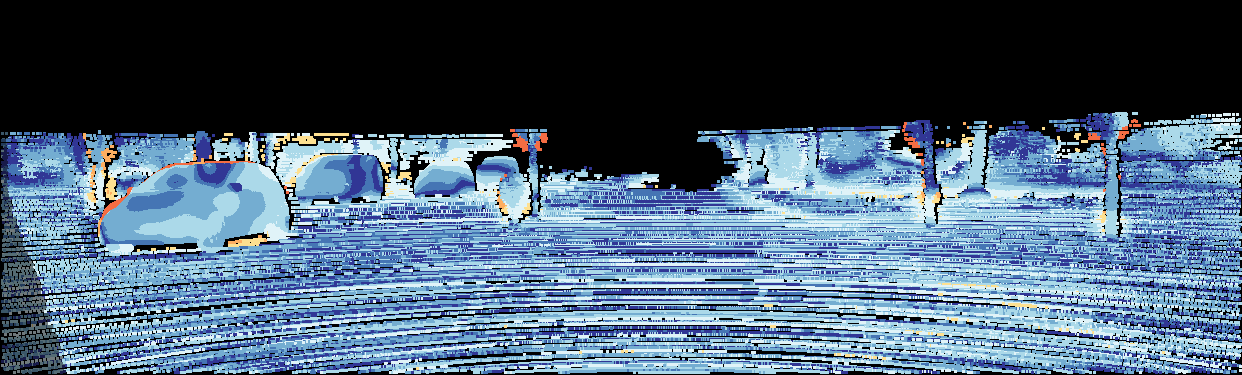}
	}
	\\
	\vspace{-0.7cm}
	\hspace{-0.2cm}
	\subfloat[]{\includegraphics[width=5.5cm,height=1.5cm]{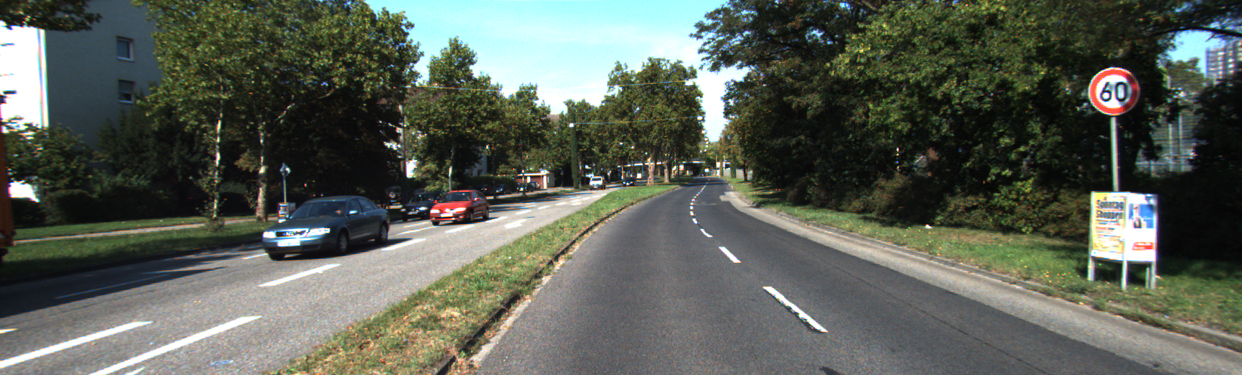}
	}
	\hspace{0cm}
	\subfloat[]{\includegraphics[width=5.5cm,height=1.5cm]{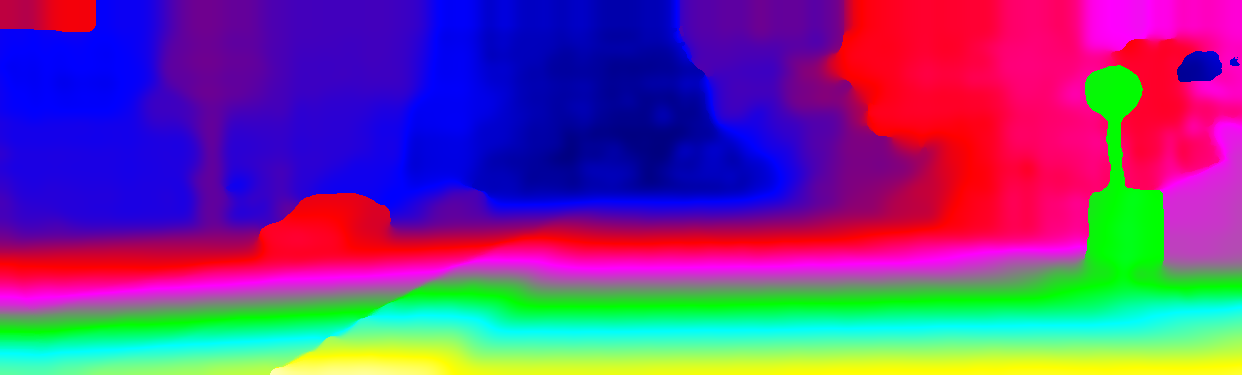}
	}
	\hspace{0cm}
	\subfloat[]{\includegraphics[width=5.5cm,height=1.5cm]{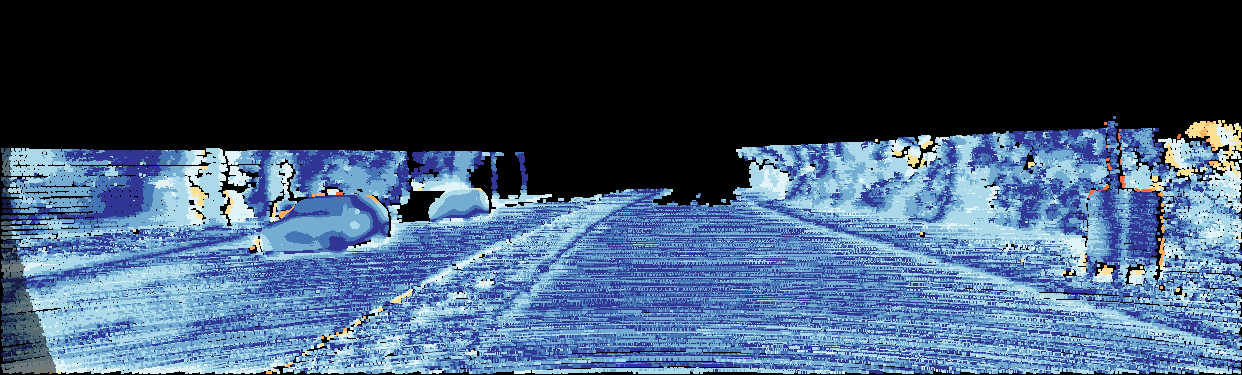}
	}
	\\
	\vspace{-0.5cm}
	\caption{\small Qualitative results of: (left) input images, (center) predicted disparity maps, and (right) stereo errors.}  
	\vspace{-0.5cm}
	\label{fig:QualitativeKITTI}
\end{figure*}
 \begin{table}
	\small
	\setlength{\tabcolsep}{2.5pt}
	\renewcommand{\arraystretch}{1}
	\centering
	\caption{\small Stereo matching results (end point error and pixel percentages with errors larger than 2, 3 and 5 pixels) of different ablation versions of the LRCR model on the KITTI 2015 validation set. The results of different steps for the recurrent versions are shown.}
	\vspace{-0.2cm}
	\begin{tabular}{c||c|c|c|c|c}
		
	   Model	& Recurrent Type & \textgreater 2 px& \textgreater 3 px & \textgreater 5 px & EPE  \\
		
		\hline
		WTA & -- &7.94 & 4.78 & 3.11 & 1.144  \\
		\hline
		Non-recurrent & -- & 5.70 & 3.69 & 2.44 & 0.939    \\ 
		\hline
		\multirow{3}{*}{$t$=1}  & w/o comp & 6.24 & 3.92 & 2.54 & 0.972   \\
		& w/o atten & 6.15 &  3.87 & 2.53 & 0.969   \\
		& LRCR  & 6.20 & 3.90 & 2.54 & 0.971 \\ \hline
		\multirow{3}{*}{$t$=2}  & w/o comp & 5.88 & 3.74  & 2.46 & 0.950\\
		& w/o atten & 5.79& 3.72& 2.45 & 0.945\\
		& LRCR & 5.61 & 3.66 &2.42 &0.935 \\   \hline
		\multirow{3}{*}{$t$=3} & w/o comp & 5.64 & 3.67 & 2.43& 0.936\\
		& w/o atten & 5.05& 3.44& 2.29 & 0.891 \\
		& LRCR & 4.85 & 3.35 & 2.15 & 0.877   \\ \hline
		\multirow{3}{*}{$t$=4} 	& w/o comp & 5.43& 3.61& 2.38& 0.922\\
		& w/o atten & 4.54& 3.22& 2.08& 0.855\\  
		& LRCR  & 4.33 & 3.12  & 2.02  & 0.839   \\  \hline
		\multirow{3}{*}{$t$=5} 	& w/o comp & 5.32 & 3.58& 2.36& 0.913\\
		& w/o atten & 4.14& 3.05& 1.99& 0.825\\  	                        
		&LRCR    & 3.92 & 2.96  & 1.93  & 0.806   \\
		\hline
	\end{tabular}
	\label{tab:KITTI1}%
	\vspace{-0.6cm}
\end{table}

 \paragraph{Results.} Based on the matching cost volumes produced by the constant highway network, we may directly apply the ``Winner Takes All'' (WTA) strategy, or choose different ablation versions of the LRCR model to generate the disparity maps. We thus compare  WTA,  full LRCR, and its multiple variants including non-recurrent, non-comparison, and non-attention. The non-recurrent version is a one-step feed-forward network explained in Sec.~\ref{sec:training}. The non-comparison version does not perform left-right comparison at each recurrent step. Instead, it just updates the left-view disparity based on only the left-view matching cost volumes at each step. The non-attention version does not include the left-right comparative branch to learn the attention map. Instead, it replaces the learned attention maps with a simple subtraction on the two disparity maps (\emph{e.g.}, $D_{\text{left}}$ and $D_{\text{left}}^{\prime}$). 
  \begin{table}
	\small
	\newcommand{\tabincell}[2]{\begin{tabular}{@{}#1@{}}#2\end{tabular}}
	\setlength{\tabcolsep}{2.5pt}
	\renewcommand{\arraystretch}{1.1}
	\centering
	\caption{\small Stereo matching results (error rates of the non-occluded pixels and all the pixels) of the LRCR model and the state-of-the-art methods on the KITTI 2015 testing set.}
	\begin{tabular}{|c|c|ccc|}
		\hline
		Type	&Method & NOC & ALL &   Runtime   \\ \hline
		\multirow{4}{*}{Others}&	MC-CNN-acrt~\cite{zbontar2015computing} & 3.33 & 3.89 & 67s \\ 
		&	Content-CNN~\cite{luo2016efficient}& 4.00 & 4.54 &1s \\ 
		&	Displets v2~\cite{Guney2015CVPR}& 3.09 & 3.43 &265s \\ 
		&	DRR~\cite{gidaris2016detect}& 2.76 & 3.16 & 1.4s \\ \hline
		\multirow{2}{*}{\tabincell{c}{End-to-end\\ CNN}}	& GC-NET~\cite{kendall2017end} & 2.61 & 2.87 & 0.9s \\ 
		&	CRL~\cite{pang2017cascade}  & 2.45 & 2.67 &  0.47s \\ \hline
		\multirow{4}{*}{\tabincell{c}{LRCR\\ and the baseline}} &	$\lambda$-ResMatch (fast)~\cite{shaked2016improved}& 3.29 & 3.78 &2.8s \\
		&	$\lambda$-ResMatch (hybrid)~\cite{shaked2016improved} & 2.91 & 3.42 &48s \\ 
		&Ours (fast) &2.79 &3.31 & 4s\\ 
		&Ours (hybrid) & 2.55 & 3.03 &  49.2s\\ \hline

	\end{tabular}
	\label{tab:KITTI2}%
	\vspace{-0.4cm}
\end{table}

 Table~\ref{tab:KITTI1} shows the results of different versions of the LRCR model on the KITTI 2015 validation set. It can be seen that ``WTA'' performs worse than all the deep network based methods. All the recurrent versions achieve better results in the final step than the non-recurrent models. The results of the early steps (\emph{e.g.}, 1st or 2nd steps) of the recurrent versions are slightly worse than the non-recurrent version. The reason may be that the recurrent versions are optimized at all the steps and the gradients of the later steps are propagated back to the early steps, making the prediction of early steps less accurate. Among the three recurrent versions, the full LRCR model consistently performs the best at all the recurrent steps. The non-comparison version produces the worst results without the guiding information from the opposite view.
 
 To better demonstrate the effectiveness of the proposed LRCR model, we illustrate the learned attention maps and the real error maps at different recurrent steps in Fig.~\ref{fig:comp}. The real error maps are the difference maps between the predicted and  ground-truth disparity maps. As can be seen, the learned attention maps have high consistency with the real error maps at different steps.  Both the learned attention maps and the real error maps mainly highlight the error-prone regions, including reflective regions, occluded objects, and sophisticated boundaries. The consistency between the erroneously predicted regions and the left-right mismatched regions  is crucial for the designing of the LRCR model. This makes  it possible that the left-right comparative branch can  localize the erroneously labeled pixels with only the predicted disparities from  both views, without  explicit supervision over the erroneously labeled regions.  With more recurrent steps, the learned attention maps become less highlighted, indicating the increasing consistency of the disparity maps from the two views. We also show the qualitative visualization of the final  predicted disparity maps by the LRCR model on the KITTI 2015 testing set in Fig.~\ref{fig:QualitativeKITTI}.
 \begin{table}
 	\small
 	\setlength{\tabcolsep}{2.5pt}
 	\renewcommand{\arraystretch}{1.1}
 	\centering
 	\caption{\small Stereo matching results (end point error and pixel percentages with errors larger than 1, 3 and 5 pixels) of different variants of the LRCR model, and the state-of-the-art methods on the Scene Flow  testing set.}
 	\vspace{-0.2cm}
 	\begin{tabular}{c||c|c|c|c}
 		
 		Model & \textgreater 1 px& \textgreater 3 px & \textgreater 5 px & EPE  \\
 		\hline
 		WTA & 29.68 & 18.47 & 12.16& 8.07  \\
 		\hline
 		GC-NET \cite{kendall2017end} &16.9 &9.34 &7.22 & 2.51 \\ 
 		\hline
 		Non-recurrent  & 23.32 & 14.08 & 9.69 & 5.26    \\ 
 		\hline
 		LRCR (t=1)    & 25.61 & 15.53 & 10.57 & 6.33   \\
 		LRCR (t=2)    & 21.36 & 12.80 & 9.03 & 4.54   \\
 		LRCR (t=3)    & 18.59 & 10.84 & 7.86 & 3.18   \\
 		LRCR (t=4)    & 16.74 & 9.54  & 7.03  & 2.43   \\
 		LRCR (t=5)    & 15.63 & 8.67  & 6.56  & 2.02   \\
 		\hline
 	\end{tabular}
 	\label{tab:SF}%
 	\vspace{-0.2cm}
 \end{table}
 
 \begin{table}
  	\small
 	\setlength{\tabcolsep}{2.5pt}
 	\renewcommand{\arraystretch}{1}
 	\centering
 	\caption{\small Stereo matching results (error rates) on the Middlebury testing set.}
 	\vspace{-0.5cm}
 	\begin{tabular}{|c|c|}
 		\hline
 		Method & Error Rate  \\
 		\hline
 		MC-CNN-acrt\cite{zbontar2015computing} & 8.08 \\
 		$\lambda$-ResMatch (hybrid)\cite{shaked2016improved} & 9.08 \\
 		LRCR (hybrid) & 7.43 \\
 		\hline
 	\end{tabular}
 	\label{tab:Middle}%
 	\vspace{-0.4cm}
 \end{table}

 We then compare the LRCR model against the state-of-the-arts  on the KITTI 2015 testing set. The top-ranked methods of the online leaderboard is shown in Table~\ref{tab:KITTI2}. Since our model is based on the matching cost volumes obtained by~\cite{shaked2016improved}, we show the comparisons to \cite{shaked2016improved} based on both the fast and hybrid versions. From Table \ref{tab:KITTI2}, the proposed LRCR model outperforms \cite{shaked2016improved} significantly while has similar running time to \cite{shaked2016improved}. It is worth mentioning that the main time consumption lies in the matching cost computation at all possible disparities, especially when fully-connected layers are used in the matching cost generation. 
 
 End-to-end models may have slightly better results than the LRCR model, but they are all trained with a huge amount of extra training data \cite{mayer2016large}. For example, \cite{kendall2017end} and \cite{pang2017cascade} utilize the large Scene Flow training set to train their models (\emph{i.e.}, extra 34k pairs of training samples for GC-NET \cite{kendall2017end} and extra 22k pairs of training samples for CRL \cite{pang2017cascade}). The demand for the huge training set is due to the very deep networks that they use to achieve the end-to-end disparity estimation. For example, the CRL model \cite{pang2017cascade} contains 47 convolutional layers with around 74.95M learnable parameters in its two-stage cascaded architecture. In comparison, our LRCR model has only about 30M learnable parameters, including both the hybrid version of the constant highway networks for  matching cost learning and the stacked ConvLSTMs for disparity estimation. We also claim the possibility of the LRCR model in achieving better disparity results when leveraging stronger networks to learn the matching costs, \emph{e.g.}, embedding the LRCR model to the existing end-to-end networks.
 
 \subsection{Experiments on Scene Flow}
 \label{sec:experimentSCENEFLOW}
\paragraph{Dataset.} The Scene Flow dataset~\cite{mayer2016large} is a large scale synthetic dataset containing around 34k stereo pairs for training and 4k for testing.
\vspace{-0.2cm}
\paragraph{Implementation details.} First, the hybrid version of the constant highway network~\cite{shaked2016improved} is used to learn the matching cost volumes. $d_{\mathrm{max}}$ is set to 320. Each parallel ConvLSTM in the LRCR model has four ConvLSTM layers with $3*3$ convolutions. Three additional $1*1$ convolutional layers are also added after the ConvLSTM layers. The numbers of output channels in the four ConvLSTM layers and the three $1*1$ convolutional layers are $d_{\mathrm{max}}$, $2d_{\mathrm{max}}$, $2d_{\mathrm{max}}$, $d_{\mathrm{max}}$, $d_{\mathrm{max}}$, $d_{\mathrm{max}}$, and $d_{\mathrm{max}}$, respectively. The numbers of epochs for both training stages and the learning rate setting are the same as those in the KITTI 2015 experiments.
\vspace{-0.4cm}
\paragraph{Results.} We show the Scene Flow testing set results of applying ``WTA'', the non-recurrent version model, and the full LRCR model in Table~\ref{tab:SF}. The LRCR model outperforms the state-of-the-art end-to-end GC-NET \cite{kendall2017end} noticeably. One can also find that the improvement trend on the Scene Flow dataset is similar to that on  KITTI 2015. The relative improvements between consecutive steps become smaller with the increasing of the recurrent steps. The LRCR model has better results than the non-recurrent version model at the later steps, while performs slightly worse at the 1st step. The ``WTA'' strategy is inferior to all the deep network based methods.

\vspace{-0.2cm}
\subsection{Experiments on Middlebury}
 \label{sec:experimentMIDDLEBURY}
The Middlebury dataset contains five subsets~\cite{scharstein2002middlebury} of indoor scenes. The image pairs are indoor scenes given a full, half and quarter resolutions.  Due to the small number of training images in  Middlebury, we learn the matching costs by fine-tuning the  constant highway network model in the Scene Flow on the Middlebury training set. For the LRCR model training, we directly train the full LRCR model with 5 recurrent steps by fine-tuning the LRCR model trained on the Scene Flow dataset. The training lasts for 10 epochs with the initial learning rate set to 0.001 and decreased by a factor of 10 after 5 epochs.

Table~\ref{tab:Middle} shows the result comparison between the LRCR model and \cite{shaked2016improved} which shares the same matching cost as our method. The proposed LRCR model outperforms \cite{shaked2016improved} significantly. The LRCR model achieves better results than another deep learning based method which  shows outstanding performance on this dataset, \emph{i.e.}, MC-CNN-acrt \cite{zbontar2015computing}.

\vspace{-0.3cm}
\section{Conclusion}
In this paper, we  proposed a novel left-right comparative recurrent model, dubbed  LRCR, which is capable of performing left-right consistency  check and   disparity estimation jointly. We also introduced a soft attention mechanism  for better guiding the model itself to selectively focus on the unreliable regions for subsequent refinement. In this way, the proposed LRCR model is shown to  progressively improve  the disparity map estimation.  We conducted extensive evaluations on the KITTI 2015, Scene Flow and Middlebury datasets, which   validate that LRCR outperforms  conventional disparity estimation networks   with offline left-right consistency check, and achieves comparable results with the state-of-the-arts.

\vspace{-0.2cm}
\section*{Acknowledgments}{
	Jiashi Feng was partially supported by NUS startup R-263-000-C08-133,
	MOE Tier-I R-263-000-C21-112, NUS IDS R-263-000-C67-646 and ECRA R-263-000-C87-133.
}
{\small
\bibliographystyle{ieee}
\bibliography{egbib}
}

\end{document}